%% file: main.tex
\pdfoutput=1
\documentclass{article}

 \usepackage[main, final]{neurips_2026}

\usepackage[utf8]{inputenc} %
\usepackage[T1]{fontenc}    %
\usepackage{hyperref}       %
\usepackage{url}            %
\usepackage{booktabs}       %
\usepackage{amsfonts}       %
\usepackage{nicefrac}       %
\usepackage{microtype}      %
\usepackage[table]{xcolor}  %

\usepackage{subcaption}
\usepackage{wrapfig}
\usepackage{amsmath,amssymb,amsthm,mathtools}
\usepackage{enumitem}
\usepackage{graphicx}
\usepackage{algorithm}
\usepackage{algorithmic}
\usepackage{tikz}
\usepackage{adjustbox}
\usepackage[most]{tcolorbox}
\usepackage{makecell}
\usepackage{soul}              %
\usepackage{pifont}

\definecolor{darkblue}{rgb}{0, 0, 0.5}
\hypersetup{colorlinks=true, citecolor=darkblue, linkcolor=darkblue, urlcolor=darkblue,
  pdftitle={Speculative Self-Distillation enables Efficient Knowledge Internalization},
  pdfauthor={Shayan Talaei, Agam Bhatia, Arshia Soltani Moakhar, Jonas H\"ubotter, Amin Saberi, Azalia Mirhoseini}}

\title{Speculative Self-Distillation enables \\ Efficient Knowledge Internalization}

\usetikzlibrary{positioning, arrows.meta, fit, backgrounds, calc,
                decorations.pathreplacing, shapes.misc, shapes.symbols}

\definecolor{teacherDark}{RGB}{142,36,52}
\definecolor{teacherMid}{RGB}{205,95,100}
\definecolor{teacherLight}{RGB}{250,210,205}
\definecolor{teacherPale}{RGB}{253,237,234}
\definecolor{studentDark}{RGB}{32,78,138}
\definecolor{studentMid}{RGB}{85,135,200}
\definecolor{studentLight}{RGB}{205,225,248}
\definecolor{studentPale}{RGB}{235,243,253}
\definecolor{padGray}{RGB}{226,226,231}
\definecolor{padBorder}{RGB}{158,158,166}
\definecolor{vllmFill}{RGB}{248,245,239}
\definecolor{vllmBorder}{RGB}{105,95,80}
\definecolor{wrapperBorder}{RGB}{105,90,95}
\definecolor{wrapperFill}{RGB}{251,250,247}
\definecolor{neutralBox}{RGB}{251,251,253}

\theoremstyle{definition}
\newtheorem{definition}{Definition}
\theoremstyle{remark}

\newcommand{\E}{\mathbb{E}}

\newcommand{\KL}{D_{\mathrm{KL}}}

\newcommand{\ptheta}{\pi_\theta}
\newcommand{\pteacher}{\pi_T}

\newcommand{\cD}{\mathcal{D}}

\newcommand{\pgen}{\pi_{\mathrm{gen}}}
\newcommand{\Dswitch}{D_{\mathrm{switch}}}
\newcommand{\Dloss}{D_{\mathrm{loss}}}

\newcommand{\deltat}{\delta_t}

\newcommand{\pmix}{\pi_{\mathrm{mix}}}
\newcommand{\ssd}{\textsc{SSD}}

\newcommand{\cL}{\mathcal{L}}
\newcommand{\thetaref}{{\theta_{\mathrm{ref}}}}

\input{files/author_info}

\begin{document}

\maketitle
\ssdauthornotes

\begin{abstract}
\input{files/sections/0_abstract}
\end{abstract}

\section{Introduction}
\label{sec:introduction}
\input{files/sections/1_introduction}

\section{A Unified View of Self-Distillation}
\label{sec:setup}
\input{files/sections/2_problem_setup}

\section{Speculative Self-Distillation}
\label{sec:method}
\input{files/sections/4_method}

\section{Experiments and Results}
\label{sec:results}
\input{files/sections/5_experimental_setup}

\input{files/sections/6_results}

\section{Related Work}
\label{sec:related_work}
\input{files/sections/7_related_work}

\section{Conclusion}
\label{sec:conclusion}
\input{files/sections/9_conclusion}

\clearpage
\input{files/sections/acknowledgements}

\bibliographystyle{colm2026_conference}
\bibliography{references}

\clearpage
\appendix

\section{Limitations, Future Work, and Broader Impacts}
\label{sec:discussion}
\input{files/sections/8_discussion}

\section{Experimental Details}
\label{app:implementation}
\input{files/sections/appendix/B_implementation_details}

\section{Additional Analyses}
\label{app:analyses}
\input{files/sections/appendix/A_additional_analyses}

\section{Case Study: JSD-Routed Inference on SciKnowEval Chemistry-L3}
\label{app:jsd-routing}
\input{files/sections/appendix/C_case_study}

\newpage

\end{document}

%% file: files/author_info.tex
\author{
Shayan Talaei$^{*}$\\
Stanford University
\And
Agam Bhatia$^{*}$\\
Stanford University
\And
Arshia Soltani Moakhar$^{*}$\\
Stanford University
\AND
Jonas Hübotter\\
ETH Zürich
\And
Amin Saberi$^{\dagger}$\\
Stanford University
\And
Azalia Mirhoseini$^{\dagger}$\\
Stanford University
}
\newcommand{\ssdauthornotes}{\begingroup\renewcommand{\thefootnote}{}%
\footnotetext{$^{*}$Equal contribution.\qquad $^{\dagger}$Equal senior authorship.}%
\footnotetext{Correspondence to \texttt{\{\href{mailto:stalaei@stanford.edu}{stalaei}, \href{mailto:agam2026@stanford.edu}{agam2026}, \href{mailto:asoltan@stanford.edu}{asoltan}\}@stanford.edu}.}%
\endgroup}

%% file: files/sections/0_abstract.tex
Many real-world applications require language models to internalize knowledge that was absent from pretraining, to incorporate continually updated facts, or to learn new domain skills. Self-distillation has emerged as a natural approach for this task. A model conditioned on the source material acts as the teacher, and an unconditioned student is trained to reproduce the teacher's behavior closed-book. We propose \textbf{Speculative Self-Distillation (\ssd{})}, an efficient per-token mixed-policy method for self-distillation. The student generates each rollout by default, while the teacher intervenes only at positions where its next-token distribution substantially diverges from the student's. Unlike off-policy distillation with teacher-generated prefixes, this design keeps training rollouts close to the student's test-time behavior, while avoiding the low-signal states that fully on-policy rollouts drift into. In effect, \ssd{} lets the teacher steer the rollout exactly where the privileged context changes the prediction. On three closed-book QA benchmarks measuring fact recall, knowledge updates, and skill acquisition, \ssd{} nearly matches on-policy distillation while using $38$--$69\%$ fewer supervised tokens and consistently outperforms off-policy distillation on the held-out test set. On a forgetting stress test, \ssd{} retains the most general capability while also achieving the best in-domain accuracy and using less supervision than the on-policy baseline. Overall, \ssd{} establishes a new accuracy--efficiency--forgetting Pareto frontier for self-distillation. Project web page is available at
\href{https://ssd-distill.github.io/}{\texttt{ssd-distill.github.io}}.

%% file: files/sections/1_introduction.tex
Practical deployments often require language models to use knowledge that
was absent from pretraining, such as updated facts, new discoveries, or domain-specific information~\citep{lewis2020retrieval, petroni2021kilt,
gao2023retrieval}. A common solution is in-context learning (ICL), where the
relevant information is supplied in the prompt at inference
time~\citep{brown2020language,dong2024survey}.

While effective, this approach has several limitations: it uses
context space, requires careful prompt or retrieval design~\citep{liu2022makes}, and
adds per-query inference cost. Models can also fail to use relevant
evidence reliably even when it is present in the
context~\citep{liu2024lost,gao2023retrieval}.
Parametric learning offers a complementary approach by distilling
new knowledge into the model's weights, where it remains available
without repeatedly retrieving and processing the source material
at inference time.

A large body of work has therefore studied how to move knowledge from context into the model's weights~\citep{maini2024entigraph, lin2025activereading}. Self-distillation has emerged as one powerful tool for this setting~\citep{snell2023context_distillation, shenfeld2026sdft}. The idea is to recast the model's ICL capability as training signal, conditioning the same model on the source material to obtain a more informed token distribution from this teacher, and then distilling that distribution back into the unconditioned student. {The central difficulty is choosing which rollout states should carry that distillation signal.}

Existing self-distillation methods differ in how they generate training
rollouts, creating a tradeoff between distribution match and learning signal.
We make this tradeoff explicit with the \emph{supervision signal}: the
per-token forward Kullback--Leibler divergence between the teacher's and
student's next-token distributions along a rollout~\citep{kullback1951information}.
On-policy rollouts match the student's test-time distribution, but are
signal-starved: after the student commits to an early prefix, the teacher is
conditioned on that prefix and can only provide a coherent continuation.
Off-policy rollouts preserve stronger signal by using teacher-generated
prefixes, but train the student on trajectories it will not produce at test
time. Both choices therefore allocate supervision inefficiently. On-policy
training spends it on positions where the teacher has little to add, whereas
off-policy training spends it on states the student will not visit. This
raises our central question: can we preserve the signal of
off-policy rollouts while staying close to the student's distribution?

\begin{figure}
    \centering
    \includegraphics[width=\linewidth]{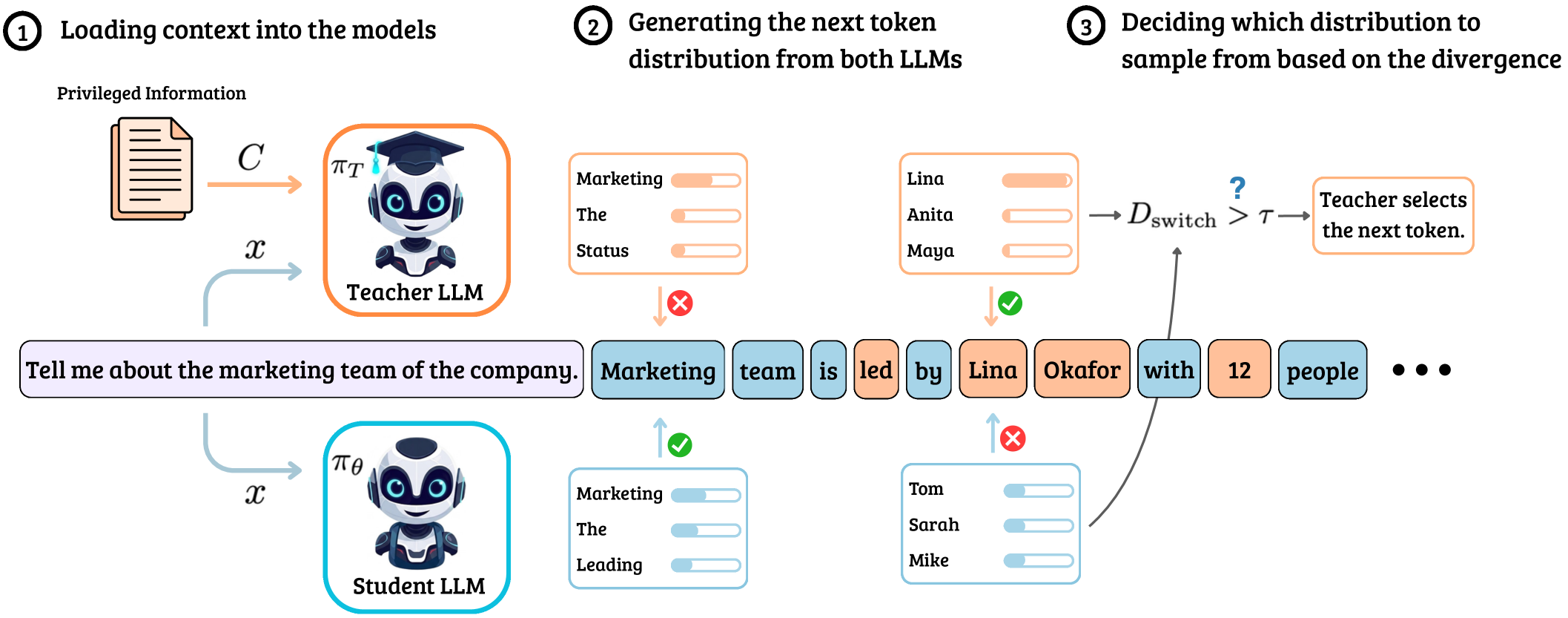}
    \caption{%
      \textbf{Speculative Self-Distillation decoding loop.}
      (1) The student receives the prompt $x$, while the teacher
      receives both $x$ and the privileged context $C$. (2) At each
      rollout position, both models propose a next-token distribution
      conditioned on the jointly decoded prefix. (3) If the switching divergence $\Dswitch$
      between the two distributions exceeds the threshold $\tau$, the
      teacher samples the next token; otherwise the student does. The
      loop continues until the rollout termination condition is met.\looseness=-1
    }
    \label{fig:method_overview}
\end{figure}

{This inefficient allocation of supervision signal motivates a method that uses
teacher generation selectively while keeping rollouts close to the student.} We propose
\textbf{Speculative Self-Distillation (\ssd{})}, a compute-efficient,
per-token mixed-policy method built around \emph{inflection tokens}: positions
where privileged context substantially changes the teacher's next-token
distribution relative to the student's. In \ssd{} the student generates the
rollout by default. At each token position, \ssd{} compares the teacher's and student's
next-token distributions. If the divergence between them exceeds a threshold $\tau$
(we use the Jensen--Shannon divergence~\citep[JSD;][]{lin1991divergence} in most experiments),
the position is marked as an inflection token and the teacher generates the next
token (Figure~\ref{fig:method_overview}). {The teacher distribution and distillation loss are still evaluated at
every response position; only the policy that emits the rollout token changes.}
The threshold $\tau$ provides a
single knob that interpolates between prior approaches: small $\tau$ recovers
off-policy distillation, while large $\tau$ recovers on-policy distillation
(Section~\ref{subsec:spectrum}).

Across three closed-book QA benchmarks, \ssd{} nearly matches on-policy accuracy while using substantially fewer supervised tokens, reducing supervised-token usage by $48\%$, $38\%$, and $69\%$ relative to on-policy self-distillation on Company Memo, Wikipedia, and Science Q\&A, respectively. \ssd{} is $1.35$ points below on-policy on Company Memo and $0.08$ and $0.15$ points above it on Wikipedia and Science Q\&A. Across 4B, 7B, and 14B models on Science Q\&A, \ssd{} achieves at least on-policy accuracy with $23$--$55\%$ fewer supervised tokens at the evaluations reported in Table~\ref{tab:scale_light}. Crucially, this efficiency does not come at the expense of the model's prior capabilities. On a forgetting stress test that fine-tunes Qwen2.5-7B on a chemistry split, we measure the mean accuracy change ($\Delta$) across six broad benchmarks relative to the base model. In this setting, \ssd{} forgets less than either baseline while also achieving the highest in-domain accuracy. Its average general performance drops by $1.4$ points, compared with $4.5$ points for on-policy and $6.6$ points for off-policy distillation. Compared with on-policy self-distillation, \ssd{} therefore improves all three axes of accuracy, forgetting, and efficiency at once in this experiment, rather than trading one against another. Our contributions are as follows:

\begin{itemize}[leftmargin=1.5em, itemsep=2pt]
  \item \textbf{A signal-based diagnosis of the off-policy/on-policy tradeoff.}
    Using the per-token teacher--student divergence as a diagnostic for the
    supervision available along a rollout, we show that on-policy
    self-distillation is signal-starved at later positions while off-policy
    rollouts carry signal throughout, accounting for on-policy training's
    compute inefficiency (Section~\ref{sec:setup:signal}).
  \item \textbf{Speculative Self-Distillation.}
    We propose \ssd{}, a per-token mixed-policy method in which the
    student generates by default and the teacher takes over only at
    \emph{inflection positions} where the teacher--student divergence
    exceeds a threshold $\tau$. A single $\tau$ interpolates between
    off-policy ($\tau\!\to\!0$) and on-policy ($\tau\!\to\!\infty$)
    self-distillation (Section~\ref{sec:method}).
  \item \textbf{On-policy accuracy at a fraction of the supervision.}
    Across three knowledge-acquisition tasks (private documents, Wikipedia
    knowledge updates, and scientific Q\&A), \ssd{} nearly matches on-policy
    self-distillation accuracy using $38$--$69\%$ fewer supervised
    tokens. We also show that a sweep over $\tau$ traces the accuracy--efficiency Pareto frontier
    between the off-policy and on-policy endpoints
    (Section~\ref{sec:results}).
  \item \textbf{Reduced forgetting without sacrificing in-domain accuracy.}
    On a forgetting stress test, \ssd{} retains the most general capability while achieving the highest
    in-domain accuracy and using less supervision than the on-policy baseline (Section~\ref{subsec:forgetting}).

\end{itemize}

%% file: files/sections/2_problem_setup.tex
Let \(\cD\) be a distribution over pairs \((C,x)\), where \(x\) is the prompt and \(C\) is the privileged context: a document, instruction,
reference answer, or other information useful for answering \(x\).
Self-distillation defines a student \(\ptheta(\cdot \mid x)\) and teacher $\pteacher(\cdot \mid x, C) := \pi_\thetaref(\cdot \mid x, C)$, which has the privileged information $C$ in context and uses the initial model weights $\theta_\mathrm{ref}$.
 The goal of \emph{self-distillation} is to update \(\theta\) so that \(\ptheta\) approximates \(\pteacher\) for held-out pairs \((C', x')\), without \(C'\) being supplied at test time~\citep{snell2023context_distillation}.

\paragraph{The self-distillation loss.}
The methods we study use the same token-level objective below and differ only in
how rollouts are generated:
\begin{equation}
  \cL(\theta) \;=\; \E_{y \sim \boldsymbol{\pgen}}\!\left[
    \frac{1}{|y|} \sum_{t} \Dloss\!\bigl(
      \pteacher(\cdot \mid x, C, y_{<t}) \,\big\|\, \ptheta(\cdot \mid x, y_{<t})
    \bigr)
  \right],
  \label{eq:unified_loss}
\end{equation}
where \(\pgen\) is the rollout policy and \(\Dloss\) is a divergence between
the teacher and student next-token distributions. We use the forward KL
divergence, \(\KL(\pteacher \,\|\, \ptheta)\), in our experiments. The choice of \(\pgen\) defines different methods.

\paragraph{Off-policy self-distillation.}
Setting \(\pgen=\pteacher\) gives off-policy self-distillation. The teacher
generates the response, and the student is trained to match the teacher along
teacher-generated trajectories~\citep{snell2023context_distillation,
kujanpaa2025prompt_distillation,eyuboglu2025cartridges}.
This approach is simple and data-efficient: every training position lies on a
{teacher-generated trajectory with privileged-context guidance. However,
during training the student is supervised on teacher-induced states; at test
time it must act on states induced by its own generations
\citep{ross2011dagger}.}\looseness=-1

\paragraph{On-policy self-distillation.}
Setting \(\pgen=\ptheta\) gives on-policy self-distillation. The student
generates the response, while the teacher provides token-level supervision
along the resulting student trajectory via
Eq.~\eqref{eq:unified_loss}~\citep{agarwal2024gkd,shenfeld2026sdft}. This removes the train-test
mismatch.

\paragraph{Dense supervision versus rollout intervention.}
{Throughout this paper, every response position contributes the
distillation loss in Eq.~\eqref{eq:unified_loss}, regardless of whether the
teacher or student emitted the corresponding rollout token. The rollout policy
therefore controls the \emph{states on which dense supervision is applied}; it
does not decide whether the teacher is evaluated or whether the loss is
present. This distinction is central to interpreting both the method and its
efficiency.}

\begin{figure}[t]
  \centering
  \includegraphics[width=\linewidth]{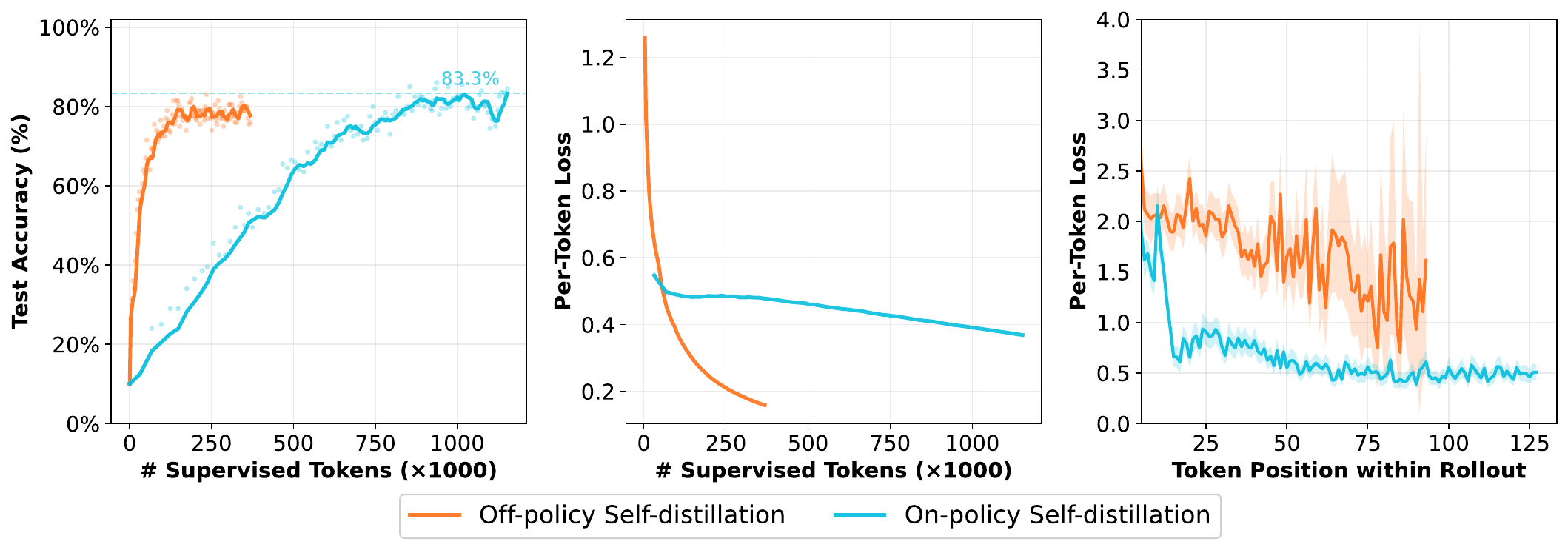}
  \caption{\textbf{Comparison of off-policy and on-policy training} on the Wikipedia task described in Section~\ref{sec:results}. \textbf{Left:}~Off-policy training converges faster than on-policy training, but saturates at lower performance. \textbf{Middle:}~Off-policy training has a larger initial self-distillation loss than on-policy training and converges faster. \textbf{Right:}~We measure the average per-token self-distillation loss across different token positions within a rollout. We find that a teacher conditioned on student-sampled prefixes predicts much like the student itself. This reduces the supervision signal at later positions of the rollout. Shaded bands (right panel) are 90\% bootstrap intervals of the mean.}
  \label{fig:signal}
\end{figure}

%% file: files/sections/4_method.tex
We propose an alternative variant of self-distillation, called \emph{speculative self-distillation} (\ssd{}), that is more efficient than on-policy training while achieving better performance than off-policy training.
We first study the tradeoff between off-policy and on-policy training in Section~\ref{sec:setup:signal} before introducing \ssd{} in Section~\ref{sec:method:description}.

\subsection{The Inefficiency of On-Policy Self-Distillation}
\label{sec:setup:signal}

To study the tradeoff between off-policy and on-policy self-distillation, we
compare them on the \emph{Wikipedia} knowledge acquisition task described in
Section~\ref{sec:results}. Figure~\ref{fig:signal}~(left) shows a clear
pattern: off-policy training reaches $78\%$ accuracy with roughly $5\times$
fewer supervised tokens than on-policy training, while on-policy training
reaches better final performance (run details in Appendix~\ref{app:signal_runs}). This is consistent with prior
work on on-policy distillation~\citep{agarwal2024gkd,shenfeld2026sdft}.
The usual explanation is that on-policy training avoids train-test mismatch.
We focus on the complementary question: \emph{why is off-policy training so much
more compute-efficient?}

\noindent\textbf{1) Off-policy training gives stronger supervision.}
The loss in Eq.~\eqref{eq:unified_loss} measures how much the teacher and
student disagree at a training state, and it also sets the size of the update.
We refer to it as the supervision signal. A larger signal means a larger update,
though not necessarily a more useful one (see point 2 below). Figure~\ref{fig:signal}~(middle) shows that
off-policy training starts with a substantially larger signal than on-policy
training and also reduces it at a faster rate.

\begin{figure}[H]
    \centering
   \includegraphics[width=\linewidth]{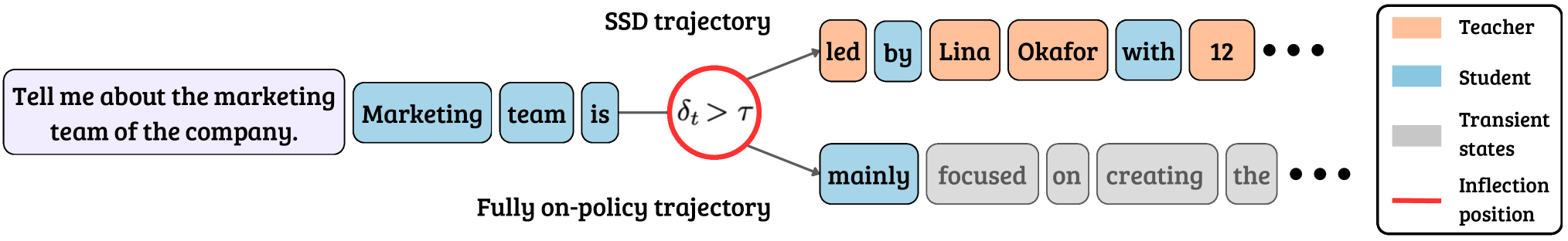}
    \caption{\textbf{Example rollout with \ssd{} versus on-policy sampling.} Both rollouts share the prefix ``Marketing team is''. At the next token the teacher and student disagree strongly, which makes it an inflection position (circled in red). \ssd{} (top) lets the teacher write this token and a few later ones (orange), and the student writes the rest (blue). The on-policy rollout (bottom) keeps the student's token, and the tokens that follow (gray) are transient states that future rollouts are unlikely to revisit.}
    \label{fig:influence_positions}
\end{figure}

\noindent\textbf{2) On-policy training spends supervision on transient states.}
Figure~\ref{fig:influence_positions} illustrates a rollout containing several
positions where the teacher and student strongly disagree.
The prompt asks about the ``marketing team of the company''.
While the first few response tokens are plausible under both student and teacher, rollouts of student and teacher diverge at the fourth token.
The teacher begins to discuss the leadership of the marketing team which is described in its privileged context whereas the student provides a generic description.
We call such a position where student and teacher disagree strongly an \emph{inflection position}: \begin{definition}[$\tau$-inflection position]
    Given a rollout $y \sim \pgen$ and $\tau > 0$, we call the $t$-th position in~$y$ a \emph{$\tau$-inflection position} if  \begin{equation}
  \deltat \;:=\; \Dswitch\!\bigl(
    \pteacher(\cdot \mid x, C, y_{<t}) \,\big\|\, \ptheta(\cdot \mid x, y_{<t})
  \bigr) > \tau,
  \label{eq:switch}
\end{equation} where \(\Dswitch\) is any divergence between the two next-token
distributions.
\end{definition}\vspace{-1ex}

At an inflection position, the distillation loss heavily pushes the student away from the original student's next-token prediction. If the student later reaches the
same prefix, it is likely to choose a different continuation.
For example, in Figure~\ref{fig:influence_positions}, the student will eventually learn to discuss the ``leadership of the marketing team''.
Several states reached by the original student rollout after
the inflection position are therefore transient: they were produced by
a policy that the update is explicitly changing. Supervision on those
states is consequently less useful.
In contrast, off-policy sampling provides detailed information about the marketing team within a single rollout.

\noindent\textbf{3) The on-policy signal decays quickly along the rollout.}
Figure~\ref{fig:signal}~(right) shows where the supervision goes. Along on-policy rollouts, the signal is high for the first few tokens and then drops, while along off-policy rollouts it stays high throughout. The student writes without the privileged context, so after a few tokens its rollout tends to drift into generic text that the context says little about. Conditioned on such a prefix, the teacher mostly agrees with the student and has little to correct. A teacher-written rollout instead stays tied to the context, so the teacher keeps disagreeing with the student at later positions. The same pattern holds on Science Q\&A (Appendix~\ref{app:signal_decay}): the signal in the first quarter of an on-policy rollout is $3.3\times$ that of the last quarter, and beyond position 512 it is $17\times$ lower than at positions 5--8. On-policy training therefore applies its loss at every position, but most of that loss falls on states where the teacher has little to teach.

\subsection{Method: Switching to the Teacher at Inflection Positions}
\label{sec:method:description}

The previous section highlights a key tradeoff: Off-policy self-distillation is
compute-efficient because teacher trajectories provide strong supervision, but
it trains the student on states it may not visit at test time. On-policy
self-distillation avoids this mismatch by sampling from the student, but much of
its supervision is weak or spent on transient states. We want the best of both:
rollouts should stay on the student's distribution when the student is already
aligned with the teacher, and switch to the teacher only where the privileged context
substantially changes the next-token distribution.

This motivates \emph{speculative self-distillation} (\ssd{}): generate from the
student, but let the teacher take over at inflection positions where the teacher and
student strongly disagree.
At an inflection position (i.e., where \(\deltat > \tau\)), the teacher generates the next token;
otherwise, the student does. The \emph{switching threshold} \(\tau\) controls how often the teacher writes the token:
lower values hand over to the teacher more often, while higher values keep rollouts
closer to the student. Training uses Eq.~\eqref{eq:unified_loss} with the mixed rollout policy
\(\pgen=\pmix\) induced by this rule. The switching divergence \(\Dswitch\), together with \(\tau\), determines which model emits each token, while \(\Dloss\) defines the distillation loss applied at every response position. In our experiments, \(\Dloss\) is the forward KL divergence from teacher to student.

{Crucially, the teacher distribution is computed at every response
position in order to evaluate the routing test, and Eq.~\eqref{eq:unified_loss}
is also applied at every response position. The switch changes only which
policy emits $y_t$. It should therefore be understood as selective
\emph{trajectory correction}, not selective supervision.}

%% file: files/sections/5_experimental_setup.tex
We first describe our experimental setup (Section~\ref{sec:experimental_setup}), then show that \ssd{} nearly matches on-policy self-distillation accuracy while using $38$--$69\%$ fewer supervised tokens (Section~\ref{subsec:headline}). We then show that these token savings shorten end-to-end training (Section~\ref{subsec:walltime}) and that \ssd{} reduces forgetting compared to both off-policy and on-policy baselines (Section~\ref{subsec:forgetting}). Finally, we examine how teacher intervention declines during training (Section~\ref{subsec:curriculum}) and how the switching threshold controls the accuracy--efficiency tradeoff (Section~\ref{subsec:spectrum}).

\subsection{Experimental Setup}\label{sec:experimental_setup}
\textbf{Tasks.}
The first task, \emph{Company Memo}, uses a synthetic internal document about a fictional company, modeling private organizational knowledge that
does not appear online. The second task, \emph{Wikipedia}, uses articles about
natural-disaster events that are available on the web but happened after the model's knowledge cutoff date. The third task, \emph{Science Q\&A}, is built from the Chemistry Level-3 subset of
SciKnowEval~\citep{feng2024sciknoweval}, which contains undergraduate-level chemistry questions requiring scientific reasoning. All three tasks are drawn from the experiments in
SDFT~\citep{shenfeld2026sdft}.

\textbf{Models.} Unless stated otherwise, the student and the teacher are both Qwen3-4B-Instruct-2507. The forgetting study in Section~\ref{subsec:forgetting} uses Qwen2.5-7B-Instruct.

\textbf{Baselines.}
The baselines span the two ends of the self-distillation spectrum.
For off-policy distillation~\citep{snell2023context_distillation, eyuboglu2025cartridges}, we evaluate SFT, which applies cross-entropy to teacher-sampled tokens, and
FKL, which applies the forward KL divergence between teacher and student. For on-policy distillation, we evaluate SDFT
\citep{shenfeld2026sdft}. Hyperparameters for every method, including the switch and threshold of \ssd{}, are selected on the held-out split (Appendix~\ref{app:hparams}).

\textbf{Evaluation.} The headline Qwen3-4B comparisons use three independent training seeds and greedy evaluation. Curves show the mean and $\pm 1$ sample standard deviation across seeds. Other experiments use the evaluation protocols specified in their captions and appendix sections.

Per-task split sizes, hyperparameter searches, prompt templates, and
other implementation details are reported in Appendix~\ref{app:implementation}.
We include a qualitative case study in Appendix~\ref{app:jsd-routing}.

%% file: files/sections/6_results.tex
\subsection{\ssd{} reaches on-policy accuracy with fewer supervised tokens}
\label{subsec:headline}

\begin{figure}[t]
  \centering
  \includegraphics[width=\linewidth]{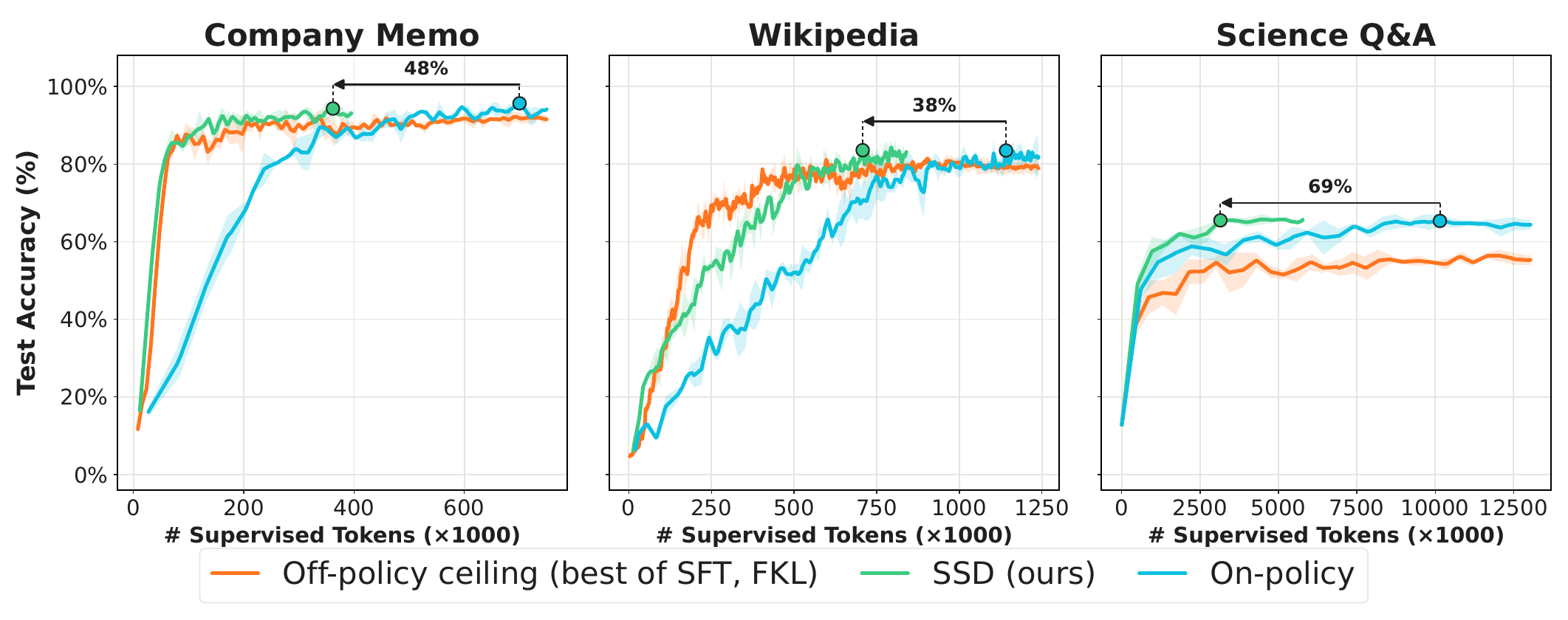}
\caption{%
    \textbf{\ssd{} approaches or exceeds the on-policy plateau with a
    fraction of the supervised tokens.}
    Test accuracy versus cumulative supervised tokens on
    Company Memo (left), Wikipedia (middle), and Science Q\&A
    (right). Each curve is the mean over three independent training
    seeds; shading is $\pm 1$ sample standard deviation. Each
    panel compares the stronger off-policy baseline (SFT or FKL), the on-policy baseline (SDFT), and \ssd{}.
    Arrows show the reduction in supervised tokens from the highest-accuracy
    on-policy point to the first \ssd{} point that reaches that accuracy.
    For Company Memo, we use \ssd{}'s highest-accuracy point instead.
}
  \label{fig:headline}
\end{figure}

\begin{table}[!t]
\caption{%
    \textbf{The token efficiency of \ssd{} persists across model
    scales on Science Q\&A.}
    Accuracy and cumulative supervised tokens for on-policy
    self-distillation and \ssd{}.
    Reported intervals are 95\% confidence intervals.
    For Qwen3-4B, we report the final evaluation after 8 epochs for
    on-policy training and 4 epochs for \ssd{}; accuracy and token counts
    are averaged over three training seeds.
    For Qwen2.5-7B, results are based on three sampled decodes.
    Qwen3-14B results are from a single run.
    The last column is the reduction in supervised tokens relative to
    on-policy training at these evaluations.
}
\centering
\begin{tabular}{lrrrrr}
\toprule
& \multicolumn{2}{c}{On-policy} & \multicolumn{2}{c}{\ssd{}} & \\
Model & Acc. & Tokens (M) & Acc. & Tokens (M) & Fewer tokens \\
\midrule
Qwen3-4B & $64.6\pm2.7$ & 13.19 & $\mathbf{66.3\pm2.6}$ & \textbf{5.88} & \textbf{55\%} \\
Qwen2.5-7B & $71.0\pm0.8$ & 2.84  & $\mathbf{73.5\pm1.7}$ & \textbf{2.20} & \textbf{23\%} \\
Qwen3-14B  & 73.2 & 2.16 & \textbf{75.5} & \textbf{1.26} & \textbf{42\%} \\
\bottomrule
\end{tabular}

\label{tab:scale_light}
\end{table}

\begin{table}[!t]
\caption{%
    \textbf{\ssd{} improves Science Q\&A accuracy while reducing
    forgetting relative to on-policy self-distillation.}
    Forgetting evaluation on Qwen2.5-7B-Instruct after fine-tuning
    on the SciKnowEval Chemistry split.
    Hella.\ = HellaSwag, Human = HumanEval (pass@1),
    TQA1 = TruthfulQA (MC1), and Wino.\ = Winogrande.
    Avg.6 is the mean score across the six general benchmarks;
    $\Delta$ is its change relative to the base model. Bold marks the best fine-tuned method (Base excluded).
    Reported intervals on SciQA indicate 95\% confidence interval
    half-widths over three sampled decodes
    (Appendix~\ref{app:forgetting}).
}
\centering
\small
\setlength{\tabcolsep}{5pt}
\newcommand{\ci}[1]{\textcolor{gray}{\scriptsize $\pm$ #1}}
\resizebox{\textwidth}{!}{%
\begin{tabular}{lcccccccccc}
\toprule
Method & Hella. & Human & IFEval & MMLU & TQA1 & Wino. & Avg.6 & $\Delta \uparrow$ & SciQA $\uparrow$ & Tokens (M) $\downarrow$ \\
\midrule
Base
& 62.1 & 70.1 & 72.9 & 71.8 & 47.9 & 71.1
& 65.98
& --
& 34.2\ci{2.5}
& -- \\

Off-policy (SFT)
& \textbf{61.0} & 64.6 & 47.0 & 69.8 & 40.9 & 72.8
& 59.34
& $-6.64$
& 65.0\ci{2.4}
& 1.82 \\

Off-policy (FKL)
& 59.6 & 64.0 & 48.4 & 69.4 & 41.1 & \textbf{73.4}
& 59.33
& $-6.65$
& 65.3\ci{3.0}
& 1.82 \\

On-policy
& 59.6 & \textbf{73.8} & 49.5 & 69.9 & 43.1 & 73.2
& 61.51
& $-4.47$
& 71.0\ci{0.8}
& 2.84 \\
\midrule

\ssd{}
& \textbf{61.0} & 70.7 & \textbf{66.8} & \textbf{71.1} & \textbf{44.7} & 73.2
& \textbf{64.59}
& $\mathbf{-1.38}$
& \textbf{73.5}\ci{1.7}
& 2.20 \\
\bottomrule
\end{tabular}%
}
\label{tab:forgetting}
\end{table}

We first evaluate whether per-token switching improves the accuracy--efficiency tradeoff relative to purely off-policy or on-policy self-distillation. Figure~\ref{fig:headline} reports test accuracy as a function of cumulative supervised tokens across our evaluation tasks. Supervised tokens count response positions that receive the distillation loss, summed over training (Appendix~\ref{app:token_accounting}). The Science Q\&A curves against training time are in Figure~\ref{fig:walltime}. The baselines exhibit a consistent tradeoff: off-policy training improves rapidly but plateaus below \ssd{} and on-policy on Wikipedia and Science Q\&A, while on Company Memo it ends close to \ssd{}. On-policy training reaches a higher plateau only after substantially more supervision. \ssd{} combines the favorable parts of both regimes. On Company Memo and Science Q\&A, it follows the steep early trajectory of off-policy training, and on Wikipedia and Science Q\&A it continues improving into the on-policy high-accuracy regime instead of flattening at the off-policy ceiling.

Relative to on-policy training, \ssd{} uses $48\%$, $38\%$, and $69\%$ fewer supervised tokens on Company Memo, Wikipedia, and Science Q\&A, with accuracy changes of $-1.35$, $+0.08$, and $+0.15$ percentage points, respectively. \ssd{} therefore stays within $1.4$ points of on-policy accuracy on every task. For both token and training-time comparisons, we use the highest accuracy on the on-policy mean curve and the first point on the \ssd{} mean curve to reach or exceed it. If \ssd{} never reaches that accuracy, as on Company Memo, we use its highest-accuracy point instead. We compute both accuracy differences and token savings from these pairs, with token savings measured relative to on-policy training.

This pattern is consistent with the central hypothesis behind \ssd{}. The useful training signal is concentrated at positions where the privileged context changes the next-token distribution, while many other positions can remain student-sampled. By intervening only at those positions, \ssd{} preserves much of the token efficiency of off-policy training while retaining the final-accuracy advantage of on-policy training.

Table~\ref{tab:scale_light} connects the Science Q\&A result to larger models. In all three models, \ssd{} achieves at least on-policy accuracy with $23$--$55\%$ fewer total supervised tokens at the reported evaluations. On the same task, closed-book \ssd{} also matches or exceeds retrieving solved demonstrations at inference time while producing shorter responses (Appendix~\ref{app:retrieval}, Table~\ref{tab:retrieval_appendix}).

\subsection{\ssd{}'s token savings translate into faster training}
\label{subsec:walltime}

\noindent\begin{minipage}[t]{0.54\textwidth}\vspace{0pt}
\hyphenpenalty=1000
Section~\ref{subsec:headline} showed that \ssd{} nearly matches on-policy accuracy with fewer supervised tokens. Whether this also shortens end-to-end training depends on the system around it: the trainer and inference engine, the hardware, and how teacher and student generation are scheduled. A naive implementation pays for an extra teacher forward pass at every decoding step. We instead modified vLLM to run the teacher and the student concurrently, with the teacher on a second GPU sending only its final hidden state to the student at each step (Appendix~\ref{app:engine}). In our training runs, this joint decoding reaches about 92\% of the throughput of student-only decoding, so a joint step takes about as long as a student step. With this setup, \ssd{} and on-policy training take nearly the same wall time per million supervised tokens (Appendix~\ref{app:timing_breakdown}). Translated into end-to-end training time on Science Q\&A, this means that \ssd{} reaches the on-policy accuracy plateau after about 130 minutes, while on-policy training needs about 414 minutes, a $69\%$ reduction (Figure~\ref{fig:walltime}). This shows that the token savings of \ssd{} carry over to faster end-to-end training.
\end{minipage}\hfill
\begin{minipage}[t]{0.44\textwidth}\vspace{-1.8em}

  \centering
  \includegraphics[width=\linewidth]{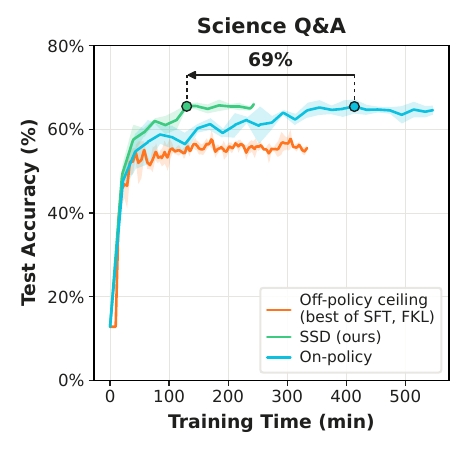}
  \captionof{figure}{\textbf{\ssd{} reaches on-policy accuracy in less training time.} Test accuracy versus training time on Science Q\&A. Curves show mean $\pm1$ SD over three independent training seeds. Evaluation time is excluded. Appendix~\ref{app:timing_breakdown} reports the training-time breakdown per supervised token.}
  \label{fig:walltime}

\end{minipage}

\subsection{\ssd{} reduces forgetting relative to on-policy methods}
\label{subsec:forgetting}

Beyond accuracy and efficiency, self-distillation should preserve the student's prior capabilities. Since on-policy self-distillation improves the accuracy--forgetting tradeoff over supervised fine-tuning~\citep{shenfeld2026sdft}, we evaluate whether \ssd{} preserves this advantage while {selectively using teacher-generated rollout tokens}. We first evaluated forgetting with Qwen3-4B, but observed little separation between methods: the six-benchmark averages remained close to the base model across all training methods, which we suspect is due to heavy post-training toward these benchmarks. To study a setting where forgetting is induced, we follow \citet{shenfeld2026sdft} and fine-tune Qwen2.5-7B on the SciKnowEval Chemistry split, measuring in-domain Science Q\&A accuracy (SciQA) against the mean accuracy change across six general benchmarks ($\Delta$). Table~\ref{tab:forgetting} reproduces the off-policy vs.\ on-policy gap: Off-policy (SFT) and Off-policy (FKL) acquire the new skill but forget the most ($\Delta \approx -6.6$), while On-policy forgets less ($\Delta=-4.47$). \ssd{} improves on this tradeoff, achieving the highest SciQA accuracy ($73.5 \pm 1.7$), the lowest forgetting ($\Delta=-1.38$), and fewer supervised tokens than On-policy ($2.20$M vs.\ $2.84$M). This suggests that mixed-policy supervision can improve the accuracy, forgetting, and efficiency tradeoff simultaneously.

\subsection{\ssd{} becomes mostly on-policy during training}
\label{subsec:curriculum}

\begin{figure}[h]
  \centering
  \includegraphics[width=\linewidth]{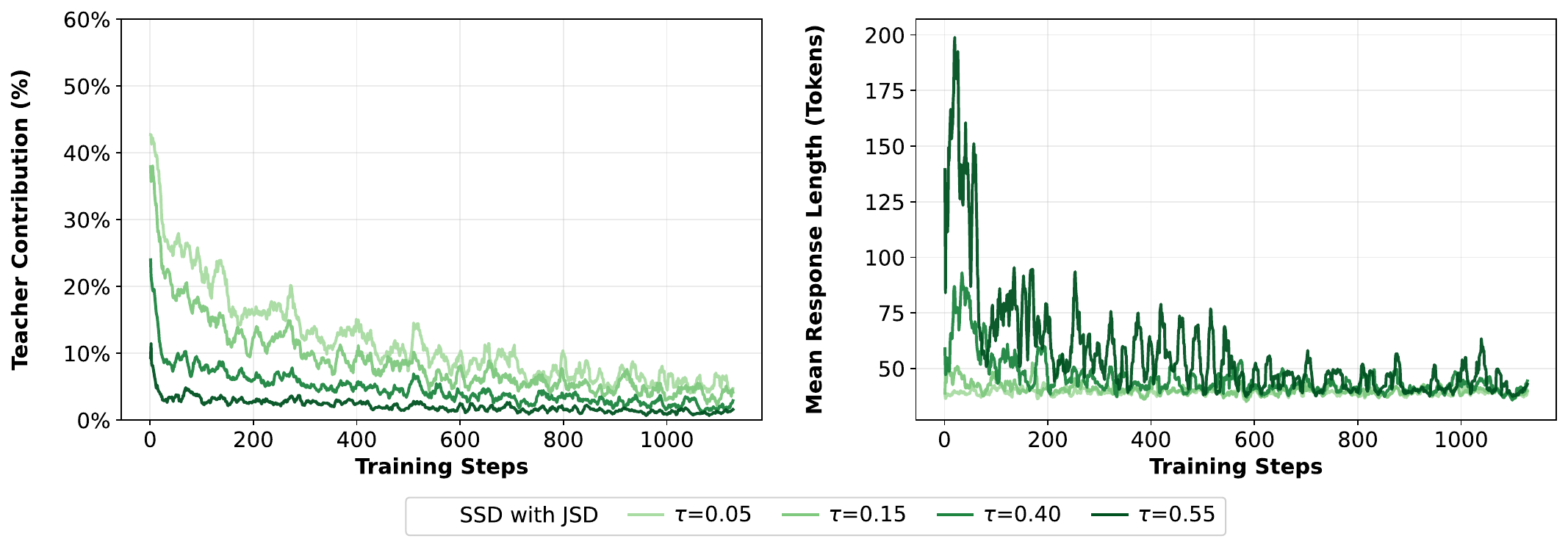}
  \caption{%
    \textbf{Teacher contribution decays over training without an external schedule.}
    Training dynamics of \ssd{} on Company Memo for four JSD thresholds $\tau$.
    \textbf{(Left)} Fraction of rollout tokens generated by the teacher, which drops from
    up to 43\% early in training to 1--5\% by the end.
    \textbf{(Right)} Mean response length, which is initially longer for higher thresholds
    and converges across thresholds as training proceeds.
  }
  \label{fig:curriculum}
\end{figure}

A desirable switching rule should not require a hand-designed schedule for how much the teacher participates. Figure~\ref{fig:curriculum} shows that \ssd{} exhibits this behavior automatically. Early in training, the teacher contributes frequently, especially at lower JSD thresholds, indicating that the student and teacher disagree at many rollout positions. As training proceeds, teacher contribution decays across all thresholds. This decline shows that the switch fires less often as the student internalizes the privileged-context behavior, shifting the rollout increasingly toward the student.

The response-length dynamics follow the same adaptation. Runs with higher thresholds initially produce longer and more variable responses, reflecting the fact that the student retains control for longer, letting its draft drift further from the teacher's distribution. As training progresses, response lengths stabilize and converge across thresholds. Thus, \ssd{} does not only reduce teacher usage; it also changes the distribution of generated trajectories as the student improves. Overall, teacher control in \ssd{} is high when the student lacks the relevant knowledge, then naturally recedes as the student becomes competent.

\noindent
\begin{minipage}[t]{0.53\textwidth}

  \subsection{\ssd{} lies on the accuracy--efficiency Pareto frontier}
  \label{subsec:spectrum}

  In Figure~\ref{fig:tau_frontier}, we evaluate whether the switching threshold $\tau$ of \ssd{} effectively interpolates between off-policy and on-policy self-distillation.
  Low thresholds $\tau \to 0$ lie near the off-policy endpoint: the teacher intervenes often, supervised-token usage is low, and final accuracy is limited. As $\tau$ increases, \ssd{} becomes more student-sampled, moving toward the on-policy endpoint and improving final accuracy.
  When using JSD as $\Dswitch$, this divergence is upper bounded by $\ln(2)\approx0.69$, and hence, any larger $\tau$ reduces to on-policy training.
  We find in Figure~\ref{fig:tau_frontier} that \ssd{} occupies a Pareto frontier: intermediate thresholds achieve higher accuracy than off-policy training while using fewer supervised tokens than on-policy training. An ablation of the switching rule itself, comparing JSD and KL switches with token-level rules from prior work, is in Appendix~\ref{app:routing_protocol}.
\end{minipage}\hfill
\begin{minipage}[t]{0.43\textwidth}
  \vspace{-6pt}
  \centering
  \includegraphics[width=\linewidth]{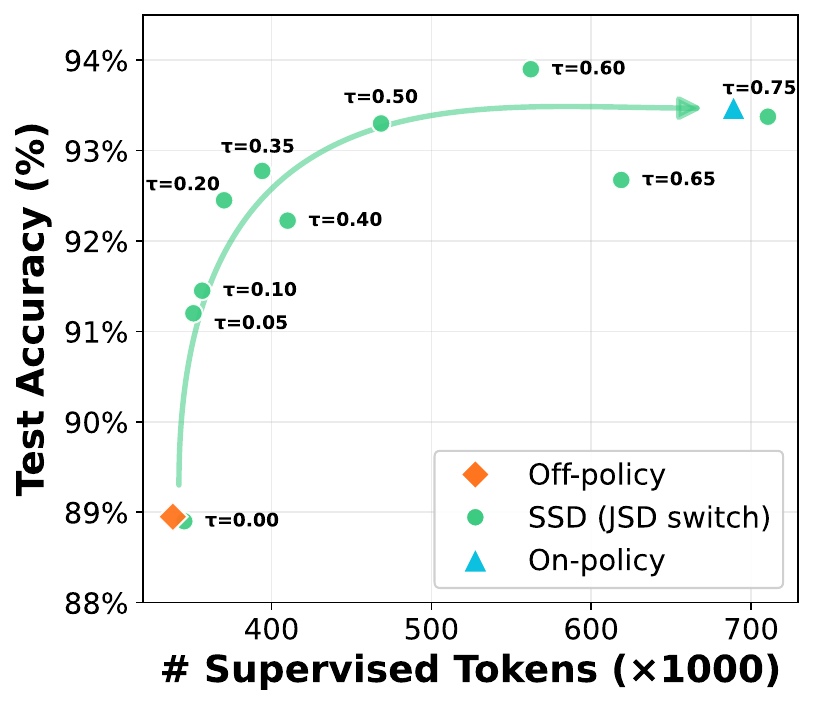}
\captionof{figure}{%
    \textbf{\ssd{} occupies the accuracy--efficiency Pareto frontier
    across switching thresholds $\tau$.}
    Downstream QA accuracy versus cumulative supervised tokens
    for Qwen3-4B on Company Memo.}
  \label{fig:tau_frontier}
\end{minipage}

%% file: files/sections/7_related_work.tex
\textbf{Knowledge distillation and rollout policies.}
Knowledge distillation trains a student model to match a teacher's outputs
\citep{hinton2015distilling} and is a standard mechanism for transferring
language-model capabilities~\citep{song2026opd_survey}. One important design
choice is how training trajectories are generated. Off-policy distillation uses
teacher-generated sequences, giving high-quality supervision but creating a
mismatch between training prefixes and the prefixes the student will produce at
test time, a form of exposure bias studied in imitation learning and sequence
prediction~\citep{ross2011dagger,bengio2015scheduled}. On-policy distillation
uses student-generated sequences and asks the teacher to supervise
them~\citep{gu2024minillm, lu2025onpolicydistillation}, reducing this mismatch
but spending teacher supervision on many tokens where the teacher provides
little new information. Mixed-policy
methods trade off these costs. GKD mixes teacher and student rollouts at the
sequence level~\citep{agarwal2024gkd}, following earlier ideas from DAgger,
scheduled sampling, and imitation knowledge distillation~\citep{ross2011dagger,
bengio2015scheduled,lin2020imitkd}. Other methods make switching decisions
within a sequence: SKD lets the teacher replace student proposals that fall outside the
teacher's top-$k$ tokens~\citep{shi2025skd}, AdaSwitch switches to the
teacher after a sliding-window KL threshold is exceeded~\citep{wu2025adaswitch},
and SWITCH applies a per-position divergence criterion~\citep{zhou2025switch}.
These works show that rollout policy is a central lever in distillation, but
they study ordinary two-model distillation; we bring this mixed-policy view to
privileged-context self-distillation.

\textbf{Context distillation and self-distillation.}
In context distillation, the teacher and student share the same base model, but
the teacher is given privileged context that the student will not receive at
test time~\citep{snell2023context_distillation}. This context may be a
document, instruction, hint, or retrieved passage, and the goal is to
internalize the resulting behavior into the model's parameters. Off-policy
variants train on responses generated with privileged
context~\citep{kujanpaa2025prompt_distillation, eyuboglu2025cartridges}, while on-policy variants supervise student-generated
trajectories using the context-conditioned model~\citep{shenfeld2026sdft,
wang2026opcd, qu2026opsd, hubotter2026sdpo, kleinebuening2026aligning,
penaloza2026privilegedinformationdistillationlanguage}. Closest to our
setting, GATES~\citep{chen2026gates} also makes privileged-context
self-distillation selective, but uses agreement among context-conditioned
tutor rollouts to gate which questions and tutor trajectories are distilled,
rather than which policy generates the rollout. Related work also
compresses context into latent state rather than model weights: Cartridges
learns compact document-specific representations, while Fast KV Compaction
constructs compact KV caches that preserve the attention behavior of the full
context~\citep{eyuboglu2025cartridges, zweiger2026fast}. Recent on-policy
self-distillation methods report better generalization to held-out questions than
off-policy training~\citep{shenfeld2026sdft}, suggesting that staying close to the student's test-time distribution is
important for robust knowledge acquisition.  However, these methods still commit
to generating the full rollout from the student, even though privileged context
typically changes the next-token distribution only at a small number of
knowledge-bearing positions. We instead ask whether the rollout policy itself
can be chosen selectively at the token level.

\textbf{Selective use of privileged information.}
Together, these two lines of work motivate keeping self-distillation trajectories close to the student while allowing teacher intervention where privileged context provides useful information. In knowledge acquisition, this information may be concentrated at a few knowledge-bearing positions, while many tokens express syntax, formatting, or reasoning shared by the teacher and student. Prior work exploits related forms of sparsity through token-level switching~\citep{shi2025skd, wu2025adaswitch, zhou2025switch}, token-level adaptation of the distillation loss~\citep{ko2025todi, xu2025selectkd, jeong2026veto}, and context compression that preserves prompt content most relevant to model predictions~\citep{zweiger2026fast}. Evidence from student-rollout distillation on unseen prompts~\citep{agarwal2024gkd, shenfeld2026sdft} and KL-based analyses of post-training behavior~\citep{shenfeld2025rls_razor} further supports keeping training close to the student's distribution. These findings motivate \ssd{}, a per-token mixed-policy method in which the student generates by default and the context-conditioned teacher intervenes when its next-token distribution differs substantially from the student's.

%% file: files/sections/9_conclusion.tex
Self-distillation for closed-book knowledge acquisition faces a tradeoff between matching the student's test-time distribution and obtaining useful supervision from teacher-generated trajectories. We introduced \ssd{}, which addresses this tradeoff token by token, letting the student generate by default and switching to the teacher at inflection positions. Across three closed-book QA tasks spanning private documents, knowledge updates, and scientific Q\&A, \ssd{} nearly matches on-policy accuracy while using $38$--$69\%$ fewer supervised tokens. On Science Q\&A, it also reaches the on-policy plateau in about $69\%$ less training time. Varying the switching threshold $\tau$ traces an accuracy--efficiency Pareto frontier between off-policy and on-policy training, while teacher intervention declines automatically as training progresses. Together with reduced forgetting, these results support adaptive token-level rollout policies for efficient knowledge acquisition.

%% file: files/sections/acknowledgements.tex
\section*{Acknowledgements and Disclosures}

We thank Hangoo Kang, Hermann Kumbong, Jacky Kwok, Yuzhen Mao, Jon Saad-Falcon, Idan Shenfeld, and Tarun Suresh for helpful discussions and feedback.

We gratefully acknowledge support from federal sources: NSF under
No.~24-554 (AIMing) and DARPA under No.~HR00112520038 (Fallingwater).
This work was supported in part by the Air Force Office of Scientific
Research (AFOSR) under Grant FA9550-23-1-0251 and in part by the Office of
Naval Research (ONR) under Grant N00014-24-1-2164.
We also gratefully acknowledge support from Stanford HAI, IBM (Code
Generation), Lightspeed, Google, and Google DeepMind. This work was also
partially supported by IBM and Felicis as an affiliate member of Stanford
Institute for Human-centered Artificial Intelligence (HAI)'s industry
program. We also gratefully acknowledge computing resources provided by
Stanford's Marlowe Computing Instrument~\citep{kapfer2025marlowe}.
The authors declare no competing interests related to this work.

%% file: files/sections/8_discussion.tex
\subsection{\ssd{} and other axes of the self-distillation pipeline}
\label{subsec:interplay}

The comparisons in this paper hold several design choices of the
self-distillation pipeline fixed so that the rollout policy is the only
variable. Each of those choices is itself an active research direction,
and how \ssd{} composes with them remains open. For example, we
treat the QA distribution $\cD$ as given, whereas
\citet{asawa2026sievesampleefficientparametriclearning}
show that document-grounded distillation becomes substantially more
sample-efficient when questions target the parts of the document the
student is most uncertain about. A more deliberate $\cD$ would reshape
the per-position signal landscape and therefore the off-/on-policy
tradeoffs.

\subsection{Self-distillation for long-horizon and reasoning tasks}
\label{subsec:reasoning}

Recent work~\citep{kim2026doesselfdistillationsometimesdegrade,
yang2026selfdistilledrlvr} has shown that on-policy self-distillation
can be less effective on reasoning-type tasks. In particular, the
information asymmetry between teacher and student means that
supervising the student to match the teacher's confidence at every
position kills the \emph{epistemic uncertainty} that the student needs
to reason its way through the rollout. We hypothesize that an \ssd{}-like
idea can help with this issue by anchoring the rollout to the teacher
only at the critical points of long sequential dependencies and
letting the student carry the rest, where a fully on-policy method
would have to learn each of those critical points one at a time. Our experiments
here are limited to closed-book QA tasks with relatively short
answers, which leaves the efficacy of \ssd{} on reasoning-heavy
tasks underexplored. Verifying it on these benchmarks is a natural
next step.

\subsection{Broader impacts}
\label{subsec:impacts}

\ssd{} lowers the cost of internalizing new knowledge into a model's weights.
As a more efficient form of self-distillation, its broader impacts are otherwise
those of the self-distillation methods it builds on.

%% file: files/sections/appendix/B_implementation_details.tex
This section gives the details needed to reproduce our experiments. We first
describe the prompt templates, the datasets, evaluation protocol and
supervised-token accounting, the hyperparameter search, and the training
infrastructure. We then give the settings of the larger-model scaling runs and
the forgetting experiment, followed by the paired teacher--student inference
engine and a breakdown of training time per supervised token.

\subsection{Prompt templates}\label{app:prompt}

The student and teacher both receive the same user question. The teacher
additionally receives a privileged context block placed immediately
before the user question. The privileged context for the teacher is the question together with
its ground-truth answer. This pairing gives the teacher the complete
and minimal information needed to answer the question, turning the
teacher into a near-oracle whose token distribution the student can
usefully imitate. The student sees only the question, so distillation
transfers the answer content into the student's weights rather than
into a retrieval prompt.

\begin{tcolorbox}[colback=black!3,colframe=black!15,boxrule=0.4pt,arc=2pt,left=6pt,right=6pt,top=4pt,bottom=4pt]
\textbf{Student prompt template}
\begin{verbatim}
[system] You are a helpful assistant. Answer the user's question.
[user]   Question: {question}
\end{verbatim}
\end{tcolorbox}

\begin{tcolorbox}[colback=black!3,colframe=black!15,boxrule=0.4pt,arc=2pt,left=6pt,right=6pt,top=4pt,bottom=4pt]
\textbf{Teacher prompt template}
\begin{verbatim}
[system] You are a helpful assistant. Answer the user's question.

         You will be provided some knowledge context relevant to the
         question. Use it when answering.
[user]   Here is context we know about the question:
         <context>
         {privileged_context}
         </context>

         Question: {question}
\end{verbatim}
\end{tcolorbox}

\subsection{Datasets, evaluation, and supervised-token accounting}\label{app:datasets}

For Company Memo and Wikipedia, we use the same set of Question-Answer
pairs released by SDFT~\citep{shenfeld2026sdft}. Company Memo is a
synthetic internal document about Hearthwell \& Sons Cooperative,
modeling private organizational knowledge that does not appear online.
For Wikipedia, we use the article on Cyclone Ditwah from the
SDFT-released set, on which the base model Qwen3-4B-Instruct-2507
scores below $20\%$, indicating that it does not possess the article's
knowledge in its weights. {For Science Q\&A, we use the Chemistry-L3 subset of}
SciKnowEval~\citep{feng2024sciknoweval}. This benchmark consists of
undergraduate-level chemistry multiple-choice questions. The model must choose
among four options and output its final answer in the format
\texttt{<answer>X</answer>}, which is evaluated against the ground-truth
choice. This setting tests whether self-distillation can internalize
domain-specific scientific knowledge and apply it to closed-book questions
that require short-form reasoning. Under the $1024$-token generation cap used in
Figure~\ref{fig:headline}, the untrained Qwen3-4B-Instruct-2507 scores
$12.8\%$ on the Science Q\&A held-out set (the origin of the Science
Q\&A curves), whereas with a $16384$-token cap it scores $49.9\%$
(Table~\ref{tab:retrieval_appendix}).
We evaluate model answers on Company Memo and Wikipedia with
Qwen3-4B-Instruct-2507 as an LLM judge, decoded greedily
(temperature $= 0$), using the same evaluation prompt as
SDFT~\citep{shenfeld2026sdft}. The judge is the base checkpoint, never a
fine-tuned one, and it is identical for every method, so differences between
methods are scored by the same fixed judge.

For the off-policy distillation methods, we augment each dataset with
teacher-generated answers, sampled from the teacher prompt template
(Section~\ref{app:prompt}) where the privileged context is the question
together with its gold answer. Targets are sampled once per run
(temperature $1.0$, top-$k=20$) and reused on every epoch. Their sampling
time is included in the wall-clock comparison in Figure~\ref{fig:walltime}.

Train and held-out split sizes per task are summarized in
Table~\ref{tab:dataset_sizes}.

\begin{table}[H]
\centering
\caption{Number of Train and Test set Question-Answer pairs per task.}
\label{tab:dataset_sizes}
\begin{tabular}{lrr}
\toprule
Task & Train & Test \\
\midrule
Company Memo            & 1133 & 100 \\
Wikipedia (Cyclone Ditwah) & 1129 & 100 \\
Science Q\&A            & 2674 & 507 \\
\bottomrule
\end{tabular}
\end{table}

\paragraph{Supervised tokens.}\label{app:token_accounting}
We count supervised tokens across all training examples and optimizer steps,
excluding prompt tokens, padding, and evaluation. For off-policy training,
a reused teacher target contributes to the count on every pass. For \ssd{},
both student- and teacher-generated response tokens count because both receive
the distillation loss.

\subsection{Hyperparameter search}\label{app:hparams}

All runs fine-tune Qwen3-4B-Instruct-2507 with AdamW, a warmup ratio of
$0.1$ and a constant learning-rate schedule (cosine on Science Q\&A).
Weight decay is $0.01$ for off-policy FKL on Science Q\&A and $0$ for the
other Science Q\&A methods.
Table~\ref{tab:hparam_sweep} gives the per-method search grids. We select
by accuracy on the task's held-out split (Table~\ref{tab:dataset_sizes}).
\ssd{} additionally selects $\Dswitch$ and $\tau$. Learning rate
$5 \times 10^{-5}$ wins for every method. Batch size $32$ wins for the
off-policy methods and $8$ for on-policy and \ssd{}; the runs in
Figure~\ref{fig:headline} use batch size $8$ for every method so that all
curves share one step size (141 optimizer steps per epoch on Company Memo
and Wikipedia, 334 on Science Q\&A). The selected switch is JSD with
$\tau=0.30$ on Company Memo and $\tau=0.50$ on Wikipedia, and reverse KL
with $\tau=1.0$ on Science Q\&A. In Figure~\ref{fig:headline}
the off-policy runs iterate on their once-sampled targets for
at least the \ssd{} training time and are shown up to the on-policy token
budget. Each selected configuration is trained with three seeds, and
Figure~\ref{fig:headline} reports the mean.

\begin{table}[H]
\centering
\caption{Hyperparameters used in our experiments. Curly braces $\{\}$
indicate a sweep over the specified values. Off-policy epoch counts are computed
from the last recorded optimizer step, with ranges across the three seeds
rounded to one decimal place. Science Q\&A runs use four epochs for \ssd{}
and eight for on-policy training.}
\label{tab:hparam_sweep}
\renewcommand{\arraystretch}{1.25}
\resizebox{\textwidth}{!}{%
\begin{tabular}{@{}llll@{}}
\toprule
\textbf{Hyperparameter} & \textbf{Off-policy SD} & \textbf{On-policy SD} & \textbf{Speculative SD} \\
\midrule
Base model           & Qwen3-4B-Instruct-2507 & Qwen3-4B-Instruct-2507 & Qwen3-4B-Instruct-2507 \\
Loss                 & CE, forward-KL         & forward-KL             & forward-KL             \\
Optimizer            & AdamW                  & AdamW                  & AdamW                  \\
LR scheduler         & \makecell[l]{Constant w.\ warmup\\(cosine on SciQA)} & \makecell[l]{Constant w.\ warmup\\(cosine on SciQA)} & \makecell[l]{Constant w.\ warmup\\(cosine on SciQA)}    \\
Warmup ratio         & $0.1$                  & $0.1$                  & $0.1$                  \\
Learning rate        & \makecell[l]{$\{2{\times}10^{-5},$ \\ $\;5{\times}10^{-5}, 10^{-4}\}$}
                     & \makecell[l]{$\{2{\times}10^{-5},$ \\ $\;5{\times}10^{-5}, 10^{-4}\}$}
                     & \makecell[l]{$\{2{\times}10^{-5},$ \\ $\;5{\times}10^{-5}, 10^{-4}\}$} \\
Effective batch size & $\{8, 32\}$            & $\{8, 32\}$            & $\{8, 32\}$            \\
Epochs               & \makecell[l]{FKL: Memo 31, Wiki 72,\\SciQA 19.9--20.4\\SFT: Memo 38, Wiki 92,\\SciQA 29.6--30.6} & 8 & Memo/Wiki 8, SciQA 4 \\
Teacher EMA rate     &                        & $\{0, 0.01\}$          & $\{0, 0.01\}$          \\
\multicolumn{4}{l}{\textit{Speculative SD-only hyperparameters}} \\
$D_{\mathrm{switch}}$ &                       &                        & \makecell[l]{reverse-KL, \\ forward-KL, \\ JSD} \\
$\tau$                &                       &                        & \makecell[l]{rev: $\{0.5, 1, 5\}$ \\ fwd: $\{0.5, 1, 5\}$ \\ JSD: $\{0.0, 0.05, \ldots, 0.65\}$} \\
\bottomrule
\end{tabular}
}%
\end{table}

\subsection{Training infrastructure and decoding}\label{app:infra}

{The experiments in Figure~\ref{fig:headline} ran on NVIDIA A100-SXM4 80\,GB GPUs: one GPU for the on-policy and off-policy baselines, and a second GPU for the resident teacher worker of \ssd{}. The remaining experiments used NVIDIA H100 80\,GB GPUs, including four GPUs per run for the Qwen3-14B scaling experiment (Appendix~\ref{app:scaling_details}).} We built our
codebase on top of the SDFT codebase
(\url{http://idanshenfeld.com/SDFT}); training is implemented on top of
the TRL training engine~\citep{trl}, and online inference for both
rollouts and judging uses vLLM~\citep{kwon2023efficient}. For training
rollouts, we set temperature $=1.0$, top-$k = 20$, and
top-$p = 1.0$ across all the baselines. Table~\ref{tab:decoding_settings}
summarizes response-length limits and evaluation decoding by experiment.

\begin{table}[H]
\centering
\small
\caption{\textbf{Response-length limits and evaluation decoding.}
Maximum new tokens excludes prompt tokens.}
\label{tab:decoding_settings}
\renewcommand{\arraystretch}{1.15}
\begin{tabular}{@{}p{0.43\linewidth}p{0.17\linewidth}p{0.32\linewidth}@{}}
\toprule
Experiment & \makecell[l]{Maximum\\new tokens} & Evaluation decoding \\
\midrule
Company Memo and Wikipedia & 512 & Greedy \\
Science Q\&A: main Qwen3-4B runs & 1024 & Greedy \\
Science Q\&A: Qwen3-14B scaling & 2048 & Greedy \\
Retrieval: base, BM25, and dense retrieval (4B and 14B) & 16384 & Greedy \\
Retrieval: \ssd{}, Qwen3-4B & 1024 & Greedy \\
Retrieval: \ssd{}, Qwen3-14B & 2048 & Greedy \\
Forgetting: Science Q\&A evaluation & 2048 & Three sampled decodes, $T=0.7$ \\
Switching-rule ablation & 2048 & Three sampled decodes, $T=0.3$ \\
Training-free JSD-routing case study & 1024 & Sampled \\
\bottomrule
\end{tabular}
\end{table}

\subsection{Larger-model scaling}
\label{app:scaling_details}

Table~\ref{tab:scale_light} asks whether the accuracy--token ordering observed
with Qwen3-4B is confined to that model size. Its rows are three separate
experiments rather than one sweep over model size. The Qwen3-14B comparison
repeats the Science Q\&A experiment with reasoning mode disabled: a single
run per method (seed 42) on four H100 GPUs with FSDP full-shard and 8-bit
AdamW, effective batch size 16, learning rate $5\times10^{-5}$, a cosine
schedule with $1.5\%$ warmup, 2 epochs (334 steps) and a frozen teacher;
\ssd{} uses a JSD switch with $\tau=0.2$. On-policy self-distillation
reaches 73.2\% after 2.16M supervised tokens, whereas \ssd{} reaches 75.5\%
after 1.26M. The Qwen2.5-7B row is the forgetting run of
Section~\ref{subsec:forgetting}, whose training settings are listed in
Table~\ref{tab:forgetting_hparams}; it differs from the Qwen3-4B runs in
schedule, epoch count and the \ssd{} learning rate.

\subsection{Forgetting experiment protocol}
\label{app:forgetting}

\paragraph{Setup.}
We fine-tune \texttt{Qwen/Qwen2.5-7B-Instruct} on the SciKnowEval Chemistry-L3 split used by \citet{shenfeld2026sdft}, consisting of 2{,}674 training examples and a 507-question held-out multiple-choice set for in-domain Science Q\&A (SciQA). Student and teacher are initialized from the same base model. We report in-domain SciQA accuracy and general-capability forgetting, measured as the mean accuracy change across six standard benchmarks relative to the base model.

\paragraph{Training.}
All methods use the shared training configuration in Table~\ref{tab:forgetting_hparams}. We sweep learning rate for all methods and, for \ssd{}, the JSD routing threshold, selecting configurations by accuracy on the held-out SciQA questions. SciQA is in-domain in the sense that these questions come from the same task as the training set; the six general benchmarks are never used for selection. The methods differ in rollout source, loss, and routing. Off-policy (SFT) trains with cross-entropy on fixed teacher-generated targets, while Off-policy (FKL) uses forward KL to the frozen teacher distribution on the same targets with softmax temperature $T=1.0$. On-policy and \ssd{} use the reasoning-prompt format, where the model emits \texttt{<reasoning>}\dots\texttt{</reasoning>} followed by \texttt{<answer>}. On-policy applies forward KL on student-sampled rollouts. \ssd{} uses mixed-policy rollouts with per-token JSD switching at threshold $\tau=0.2$.

\begin{table}[H]
\centering
\small
\caption{Shared training configuration for the forgetting experiments.}
\label{tab:forgetting_hparams}
\begin{tabular}{ll}
\toprule
Hyperparameter & Value \\
\midrule
Epochs & 2 \\
Effective batch size & 16 \\
Learning rate schedule & cosine, 1.5\% warmup \\
Gradient clipping & 1.0 \\
Precision & bf16 \\
Learning rate & $5{\times}10^{-5}$, except \ssd{} uses $3{\times}10^{-5}$ \\
Completion length & 2048 tokens for rollout-based methods \\
Loss-token mask & first 5 tokens for rollout-based methods \\
\bottomrule
\end{tabular}
\end{table}

\paragraph{Evaluation.}
We evaluate general-capability retention with the EleutherAI \texttt{lm-\allowbreak evaluation-\allowbreak harness}~\citep{eval-harness}, pinned to the SDFT-paper commit \texttt{03c44adc}~\citep{shenfeld2026sdft}, using a vLLM backend~\citep{kwon2023efficient} with \texttt{dtype=bfloat16}, \texttt{gpu\_memory\_utilization=0.85}, \texttt{max\_model\_len=8192}, and \texttt{-{}-batch\_size auto}. Table~\ref{tab:forgetting_tasks} lists the six forgetting benchmarks and metrics. We define Avg.6 as their unweighted mean and $\Delta=\text{Avg.6}-\text{Avg.6}(\text{base})$. SciQA is evaluated separately on the 507-question holdout by exact match on the extracted \texttt{<answer>} letter, generating up to 2048 tokens.

\begin{table}[H]
\centering
\small
\caption{Forgetting benchmarks and metrics.}
\label{tab:forgetting_tasks}
\begin{tabular}{lll}
\toprule
Task & Metric & Scoring \\
\midrule
HellaSwag & acc, not length-normalized & loglikelihood, deterministic \\
MMLU & acc & loglikelihood, deterministic \\
TruthfulQA & mc1 & loglikelihood, deterministic \\
Winogrande & acc & loglikelihood, deterministic \\
HumanEval & pass@1 & generative, greedy \\
IFEval & inst\_level\_loose\_acc & generative, greedy \\
\bottomrule
\end{tabular}
\end{table}

\paragraph{Confidence intervals.}
For SciQA, we report the mean and 95\% Student's $t$ confidence interval across three sampled decodes at temperature $T=0.7$.

\subsection{Paired teacher--student inference engine}
\label{app:engine}

\input{files/figures/ensemble_inference_figure}

We implemented \ssd{} with two vLLM~\citep{kwon2023efficient} workers, one for the teacher and one for the student, each on a separate GPU. A CPU controller admits batches and supplies the student prefix $x$ and the teacher prefix $[C;x]$, where $C$ is the additional context available to the teacher. Each worker maintains its own model weights and KV cache. We use vLLM's internal GPU worker and Model Runner V2 to run the models, with our own scheduling, routing, and sampling code. This adapter leaves the installed vLLM source unchanged. Figure~\ref{fig:ssd-engine} shows the inference data flow.

In our configuration, both workers prefill their prefixes, and then generate a shared continuation. At position $t$, both models condition on the same previously selected tokens $y_{<t}$. The teacher sends its final post-normalization hidden state $h^T_t$ to the student GPU through NCCL. On the student GPU, we project $h^T_t$ with a resident copy of the teacher's output head and $h^S_t$ with the student's head. We compare the logits using the switch rule and select the model and sample one token $y_t$ from the selected distribution. The student worker sends $y_t$ to the teacher through NCCL, and both workers write it into vLLM's request state on the GPU before the next forward pass.

For Qwen3-4B, sending hidden states limits the teacher-to-student transfer to $2{,}560$ values per active request and token, compared with $151{,}936$ values for a full logit vector. Routing, sampling, token exchange, and token-history updates stay on the GPUs. Keeping the selected tokens on the GPUs lets us launch the next decode step without waiting for those tokens to be copied back to the CPU. CPU workers prepare metadata and launch each decode step. They read back token IDs every eight steps. The parent controller receives the completed admission batch. With CUDA graphs enabled, we replay both vLLM decode bodies and the custom projection and routing head. We compile the divergence computation to fuse its elementwise operations and reductions. At admission, we reserve KV-cache pages for each prompt and its full continuation budget to avoid capacity-driven preemption, eviction, and re-prefill within the batch.

In the resident training configuration, the teacher worker keeps the hidden states used to predict the generated tokens on its GPU. The trainer imports them through CUDA IPC, with one device-to-device copy per admission group into its own storage. It projects the cached states with the teacher head to compute the forward-KL targets. This avoids a second teacher forward pass and storing full-vocabulary logits for the whole sequence. With sufficient GPU memory, the student's training weights, gradients, and optimizer state stay on the same GPU as the student vLLM worker, and both inference workers remain resident between rollout and training phases.

\subsection{Training time per supervised token}
\label{app:timing_breakdown}

We profile Qwen3-4B on Science Q\&A over three training seeds.
For each seed, we divide mean phase time per step by mean supervised
tokens per step, convert to minutes per million tokens, and average
across seeds (Table~\ref{tab:timing_breakdown}). Token counts follow
Appendix~\ref{app:token_accounting}, including each reuse of an off-policy
target. GPU allocations are described in Appendix~\ref{app:infra}.

\begin{table}[H]
\centering
\small
\setlength{\tabcolsep}{4pt}
\caption{Training-phase wall time in minutes per million supervised tokens
on Science Q\&A, excluding offline target generation.}
\label{tab:timing_breakdown}
\begin{tabular}{@{}lrrrrrr@{}}
\toprule
Method & Generation & \makecell[r]{Loss\\calculation} & Backward
& \makecell[r]{Update\\+ sync} & Other & Total \\
\midrule
\ssd{} & 30.25 & 3.16 & 6.37 & 0.50 & 1.10 & 41.38 \\
On-policy & 28.86 & 5.64 & 5.64 & 0.23 & 1.16 & 41.54 \\
Off-policy FKL & 0.00 & 4.01 & 6.26 & 0.19 & 0.80 & 11.25 \\
Off-policy SFT & 0.00 & 2.47 & 5.41 & 0.15 & 0.67 & 8.71 \\
\bottomrule
\end{tabular}
\end{table}

Generation includes rollout computation and bookkeeping. Loss calculation
includes training forward passes, target preparation, and loss arithmetic.
Update + sync combines gradient clipping, the optimizer update, and
measured weight synchronization to rollout workers, which we subtract
from generation. Other is the residual step time, including unassigned
data-loading, callback, and synchronization overhead. Evaluation and
checkpoint saving are excluded.

SSD and on-policy training take nearly the same wall time per supervised
token with these GPU allocations. SSD decodes with the teacher on its
second GPU concurrently with the student and reuses cached teacher hidden
states in the loss phase. The shorter loss phase offsets its higher
generation and backward times per token. Reaching matched accuracy with
fewer tokens therefore shortens training (Section~\ref{subsec:walltime}).

Off-policy training has no in-step generation. Sampling its teacher
targets once takes approximately 9.5 minutes per run, which is included
separately in Figure~\ref{fig:walltime}. Amortized over all recorded
training tokens, this adds approximately 0.33 minutes per million tokens
for FKL and 0.22 for SFT. These amortized costs depend on how many times
the targets are reused.

%% file: files/figures/ensemble_inference_figure.tex
\colorlet{ssdInk}{black}
\definecolor{ssdLine}{HTML}{BCC5CE}
\definecolor{ssdTeacher}{HTML}{FE893F}
\definecolor{ssdStudent}{HTML}{0BBFDE}
\definecolor{ssdTeacherFill}{HTML}{FEBF99}
\definecolor{ssdStudentFill}{HTML}{A5D3E7}
\colorlet{ssdTeacherRegion}{ssdTeacherFill!22!white}
\colorlet{ssdStudentRegion}{ssdStudentFill!22!white}
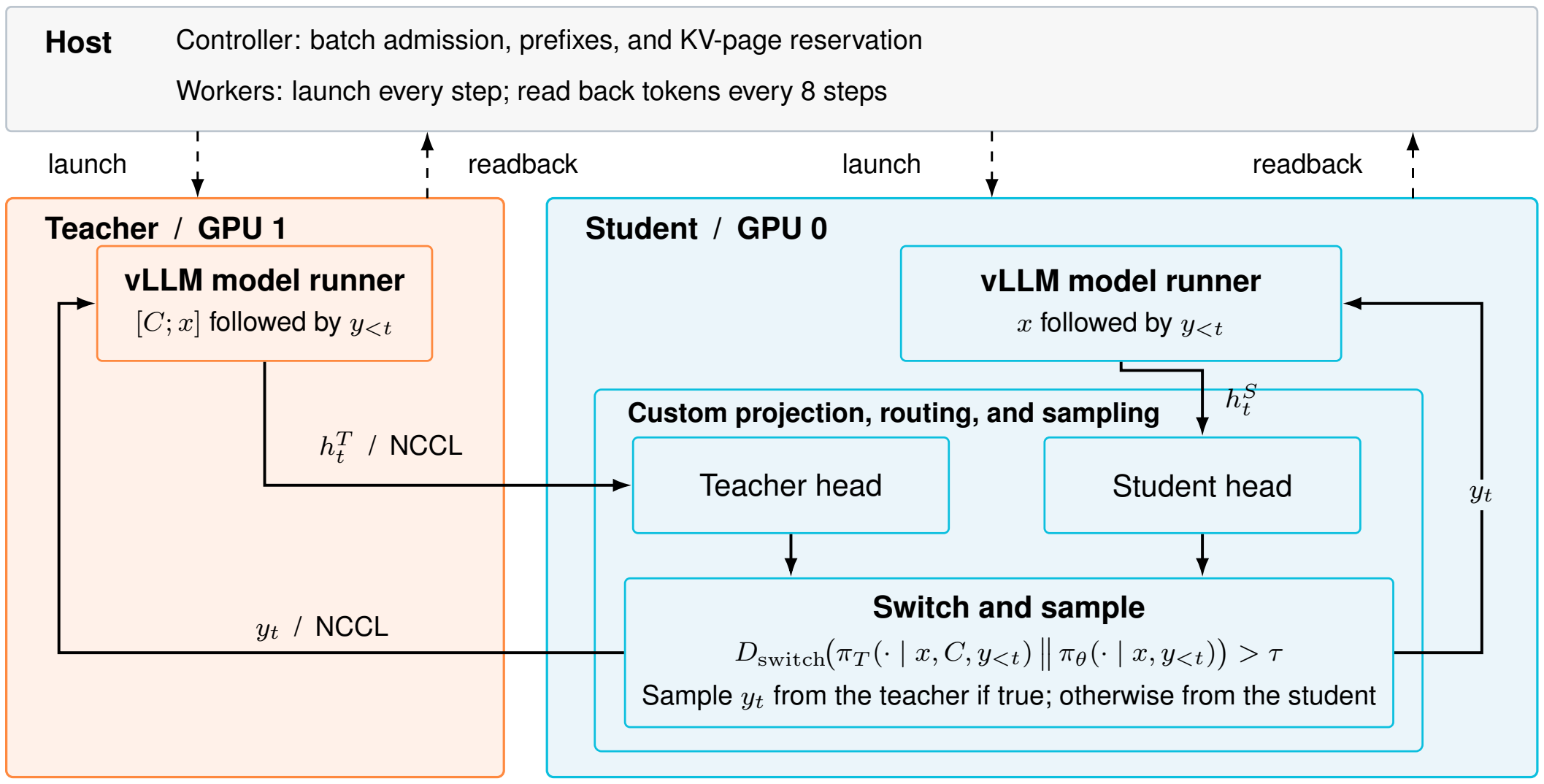
\begin{figure}[htbp]
\centering
\resizebox{\linewidth}{!}{%
\begin{tikzpicture}[
  font=\sffamily\small, text=ssdInk,
  >={Latex[length=2mm,width=1.4mm]},
  box/.style={draw=ssdLine, line width=0.6pt, rounded corners=1.5pt,
    fill=white, align=center, inner sep=5pt},
  flow/.style={->, draw=ssdInk, line width=0.85pt},
  hostflow/.style={->, draw=ssdInk, line width=0.65pt, dashed},
  label/.style={font=\sffamily\footnotesize, align=center},
  note/.style={font=\sffamily\footnotesize, text=ssdInk, align=center}
]
\node[box, fill=black!3, minimum width=160mm, minimum height=13mm]
  (host) at (8,7.55) {};
\node[anchor=west, font=\sffamily\bfseries\small] at (0.28,7.83) {Host};
\node[anchor=west, label] at (1.65,7.83)
  {Controller: batch admission, prefixes, and KV-page reservation};
\node[anchor=west, label] at (1.65,7.30)
  {Workers: launch every step; read back tokens every 8 steps};

\begin{scope}[on background layer]
  \draw[draw=ssdTeacher, fill=ssdTeacherRegion, rounded corners=2pt,
        line width=0.7pt] (0,0.15) rectangle (5.2,6.2);
  \draw[draw=ssdStudent, fill=ssdStudentRegion, rounded corners=2pt,
        line width=0.7pt] (5.65,0.15) rectangle (16,6.2);
  \draw[draw=ssdStudent, fill=ssdStudentRegion, rounded corners=2pt,
        line width=0.6pt] (6.15,0.42) rectangle (14.8,4.2);
\end{scope}
\node[anchor=west, text=ssdInk, font=\sffamily\bfseries\small]
  at (0.28,5.89) {Teacher\enspace /\enspace GPU 1};
\node[anchor=west, text=ssdInk, font=\sffamily\bfseries\small]
  at (5.93,5.89) {Student\enspace /\enspace GPU 0};

\draw[hostflow] (2.0,6.9) -- (2.0,6.2);
\draw[hostflow] (4.4,6.2) -- (4.4,6.9);
\draw[hostflow] (10.3,6.9) -- (10.3,6.2);
\draw[hostflow] (14.7,6.2) -- (14.7,6.9);
\node[note] at (0.85,6.56) {launch};
\node[note] at (5.4,6.56) {readback};
\node[note] at (9.15,6.56) {launch};
\node[note] at (13.6,6.56) {readback};

\node[box, draw=ssdTeacher, fill=ssdTeacherRegion, minimum width=35mm, minimum height=12mm]
  (teacher) at (2.7,5.1)
  {\textbf{vLLM model runner}\\[2pt]
   \footnotesize $[C;x]$ followed by $y_{<t}$};
\node[box, draw=ssdStudent, fill=ssdStudentRegion, minimum width=46mm, minimum height=12mm]
  (student) at (11.65,5.1)
  {\textbf{vLLM model runner}\\[2pt]
   \footnotesize $x$ followed by $y_{<t}$};

\node[anchor=west, font=\sffamily\bfseries\footnotesize]
  at (6.37,3.93) {Custom projection, routing, and sampling};
\node[box, draw=ssdStudent, fill=ssdStudentRegion, minimum width=33mm, minimum height=10mm]
  (thead) at (8.2,3.2)
  {Teacher head};
\node[box, draw=ssdStudent, fill=ssdStudentRegion, minimum width=33mm, minimum height=10mm]
  (shead) at (12.5,3.2)
  {Student head};
\node[box, draw=ssdStudent, fill=ssdStudentRegion,
      minimum width=80mm, minimum height=12mm]
  (route) at (10.48,1.45)
  {\textbf{Switch and sample}\\[3pt]
   \footnotesize $D_{\mathrm{switch}}\!\bigl(
     \pi_T(\cdot\mid x,C,y_{<t})\,\big\|\,
     \pi_\theta(\cdot\mid x,y_{<t})\bigr)>\tau$\\[3pt]
   \footnotesize Sample $y_t$ from the teacher if true; otherwise from the student};

\draw[flow] (teacher.south) -- (2.7,3.2) -- (thead.west)
  node[label, midway, above=3pt, xshift=-6mm, text=ssdInk] {$h^T_t$\enspace /\enspace NCCL};
\draw[flow] (student.south) -- (11.65,4.4) -| (shead.north)
  node[label, pos=0.72, right=3pt] {$h^S_t$};
\draw[flow] (thead.south) -- (thead.south |- route.north);
\draw[flow] (shead.south) -- (shead.south |- route.north);

\draw[flow] (route.west) -- (0.55,1.45) -- (0.55,5.1) -- (teacher.west);
\node[label, text=ssdInk, above] at (3.3,1.45) {$y_t$\enspace /\enspace NCCL};
\draw[flow] (route.east) -- (15.42,1.45) -- (15.42,5.1) -- (student.east);
\node[label, text=ssdInk, fill=ssdStudentRegion, inner sep=2pt] at (15.42,3.1) {$y_t$};

\end{tikzpicture}
}
\caption{\textbf{Paired vLLM inference for \ssd{}.}
The teacher and student run on separate GPUs with different prefixes and a shared continuation. At each token, the teacher sends its hidden state to the student GPU through NCCL. Both output heads, the switching test, and sampling run on the student GPU. The selected token is written into both models' request histories, with NCCL carrying it back to the teacher. CPU code launches each decode step and reads back token IDs every eight steps.}
\label{fig:ssd-engine}
\end{figure}

%% file: files/sections/appendix/A_additional_analyses.tex
These experiments probe where the supervision signal goes, how \ssd{} compares
with retrieval at inference time, and how the choice of switching rule matters.

\subsection{Supervision-signal measurements}
\label{app:signal_runs}
\label{app:signal_decay}

These experiments support the analysis in Section~\ref{sec:setup:signal}
by examining how the source of a rollout affects the supervision it provides.
We ask whether teacher-generated trajectories provide stronger feedback than
student-generated trajectories, and whether feedback weakens as a student
rollout grows longer. The Wikipedia comparison addresses both questions,
while the Science Q\&A analysis checks the positional pattern on longer
responses. In both experiments, we measure supervision with forward KL from
the privileged teacher to the student, matching the distillation loss.

On Wikipedia, Figure~\ref{fig:signal} compares single on-policy and off-policy
training runs of Qwen3-4B with a frozen teacher. Both train for eight epochs;
off-policy training reuses teacher answers sampled once per question.
Off-policy training reaches $78\%$ accuracy with roughly five times fewer
supervised tokens, while on-policy training reaches higher final accuracy.
The positional analysis uses the untrained student on all 1{,}129 training
questions, comparing teacher-written answers with greedy student rollouts
capped at 128 tokens. On student-generated prefixes, mean per-token loss
falls from roughly $1.5$--$2.0$ at early positions to about $0.5$ by position
75, while teacher-generated prefixes retain a higher signal over the plotted
range. Means include only rollouts that reach each position, and shaded
bands show 90\% bootstrap intervals over questions.

On Science Q\&A, we examine the same on-policy signal along longer responses,
using 256 rollouts from the untrained student, one per question, capped at
1024 tokens. Feedback weakens later in the response: the mean signal in the
first quartile is $3.3\times$ that in the last. Beyond position 512, it falls to about one-seventeenth of its
value at positions 5--8. As on Wikipedia, positional means include only
rollouts that reach each position.

The concentration of teacher--student disagreement early in on-policy rollouts
is consistent with the motivation for \ssd{}: as student errors accumulate,
the rollout may enter prefixes where the teacher provides weaker corrective
feedback. We therefore switch to the teacher at positions of high disagreement.
In standard on-policy training, later tokens still receive the dense
distillation loss but provide weaker feedback. \ssd{} applies the same dense
loss, using teacher interventions to keep rollouts closer to teacher-generated
trajectories and preserve stronger supervision.

\subsection{Retrieved-demonstration comparison}
\label{app:retrieval}

The main experiments ask whether self-distillation can store new knowledge in
the model's parameters so that it can answer questions without privileged
context at test time. Retrieval provides a natural alternative: keep the
knowledge external and supply related solved examples in the inference prompt.
To contextualize the closed-book results, we compare \ssd{} with retrieved
demonstrations on Science Q\&A at 4B and 14B scales.

For each evaluation question, we retrieve four training questions together
with their reference answers and insert them as solved demonstrations. BM25
uses lexical similarity, while dense retrieval uses cosine similarity under
Qwen3-Embedding-4B. We evaluate both retrieval methods, the unaugmented
closed-book base model, and the closed-book \ssd{} model using the same answer
extraction. Generated-token counts include only the response, not the retrieved
prompt.

We evaluate the final \ssd{} checkpoints. The Qwen3-4B results average the
three Science Q\&A training seeds shown in Figure~\ref{fig:walltime}; the
Qwen3-14B results use the run reported in Table~\ref{tab:scale_light}.
Generation limits are given in Appendix~\ref{app:infra}.

\begin{table}[H]
\centering
\small
\caption{\textbf{Closed-book internalization compared with retrieved
demonstrations on Science Q\&A.} BM25 and dense retrieval supply four solved
training questions; \ssd{} receives only the test question. Base and retrieval
responses are capped at $16384$ tokens; \ssd{} responses are capped at $1024$
tokens for Qwen3-4B and $2048$ tokens for Qwen3-14B. Mean generated-token
counts cover responses only, excluding the prompt.}
\label{tab:retrieval_appendix}
\begin{tabular}{llrr}
\toprule
Model & Inference method & Accuracy & Mean generated tokens \\
\midrule
Qwen3-4B & Closed-book base & 49.9 & 2864 \\
Qwen3-4B & Top-4 BM25 demonstrations & 65.9 & 2045 \\
Qwen3-4B & Top-4 dense demonstrations & 61.3 & 2146 \\
Qwen3-4B & \ssd{} closed-book & \textbf{66.3} & \textbf{508} \\
\midrule
Qwen3-14B & Closed-book base & 45.2 & 373 \\
Qwen3-14B & Top-4 BM25 demonstrations & 58.4 & 510 \\
Qwen3-14B & Top-4 dense demonstrations & 55.6 & 465 \\
Qwen3-14B & \ssd{} closed-book & \textbf{75.5} & \textbf{222} \\
\bottomrule
\end{tabular}
\end{table}

Retrieval improves the untrained model at both scales, confirming that relevant
solved examples help on this task. Nevertheless, closed-book \ssd{} matches or
exceeds the strongest retrieved-demonstration baseline while producing much
shorter responses. This comparison does not establish that parametric
internalization is generally preferable to retrieval: the retriever selects
solved training questions rather than raw source documents.

\subsection{Switching-rule ablation}
\label{app:routing_protocol}

Table~\ref{tab:routing_ablation} asks whether \ssd{} needs to compare the
full teacher and student distributions, or whether simpler token-level
signals from prior work are sufficient. To isolate the switching rule, the
ablation holds the model, data, loss, learning rate, schedule, and training
seed fixed on Science Q\&A. All ablation runs use an effective batch size of 16
and train for two epochs with a cosine learning-rate schedule and $1.5\%$ warmup.
We evaluate each run's final checkpoint with three
sampled decodes at temperature $0.3$.

We compare five rules. Following Speculative Knowledge Distillation
(SKD)~\citep{shi2025skd}, the top-$k$ rule first samples a student proposal
and lets the teacher emit the token when that proposal falls outside the
teacher's top-$k$ candidates. The entropy rule, loosely motivated by entropy-aware on-policy
distillation~\citep{xiang2026entropy_opd}, which uses teacher entropy to adapt the loss, switches the rollout on student entropy with
threshold $2.0$. The three distribution-level switches use JSD ($\tau=0.25$),
reverse KL ($\tau=1.0$), and forward KL ($\tau=0.5$). Because these
statistics have different numerical scales, we compare the rules through the
realized fraction of tokens emitted by the teacher. As in the Science Q\&A
runs of Figure~\ref{fig:headline}, the teacher in these runs is an
exponential moving average of the student with mixing rate $0.01$.

\begin{table}[H]
\caption{\textbf{Switching-rule ablation on Science Q\&A.} All rows share
the training configuration, seed, and forward-KL loss; only the switching
rule changes. Accuracy is the mean and 95\% confidence interval over three
sampled decodes. Teacher-token fraction is the fraction of rollout positions
generated by the teacher at the end of training.}
\centering
\begin{tabular}{lrrrr}
\toprule
Switching rule & Setting & Teacher-token frac. & Accuracy & Tokens (M) \\
\midrule
JSD & $\tau=0.25$ & 0.7\% & $63.3\pm3.4$ & 4.47 \\
Top-$k$ inclusion & $k=1$ & 9.3\% & $62.0\pm2.0$ & 4.51 \\
Entropy & $\tau=2.0$ & 11.5\% & $59.2\pm1.2$ & 5.18 \\
Reverse KL, $\KL(\ptheta\,\|\,\pteacher)$ & $\tau=1.0$ & 1.5\% & $65.2\pm0.3$ & \textbf{4.02} \\
Forward KL, $\KL(\pteacher\,\|\,\ptheta)$ & $\tau=0.5$ & 3.4\% & $\mathbf{67.0\pm3.1}$ & 4.07 \\
\bottomrule
\end{tabular}
\label{tab:routing_ablation}
\end{table}

For the top-$k$ rule we sweep $k\in\{20,5,1\}$. The student's proposal
already lies within the teacher's top-20 candidates at $99.7\%$ of positions
and within its top-5 at $98.5\%$, so these two settings almost never invoke
the teacher and reduce to nearly on-policy generation;
Table~\ref{tab:routing_ablation} therefore reports the non-degenerate $k=1$
setting. Even this setting and the entropy rule reach lower mean accuracy
than switching on JSD, forward KL, or reverse KL, consistent with prior work
using distribution-level disagreement for adaptive
switching~\citep{wu2025adaswitch,zhou2025switch}. These results suggest that
the context can substantially redistribute probability mass without moving
the student's proposal outside the teacher's top-$k$. The three
distribution-level switches perform similarly, with overlapping intervals,
and in this single-seed comparison we see no clear difference among them: the intervals are over three decodes of one training seed, and the rules fire at different teacher fractions ($0.7\%$ to $11.5\%$), so the rows are not matched on teacher share. We use JSD as the default because its $\ln 2$ upper bound puts $\tau$ on a fixed, interpretable scale; the Science Q\&A runs of Figure~\ref{fig:headline} use the reverse-KL switch.

%% file: files/sections/appendix/C_case_study.tex
\definecolor{teachercol}{HTML}{FDBA74}   %
\definecolor{teachertxt}{HTML}{7C2D12}   %

\sethlcolor{teachercol}
\newcommand{\T}[1]{%
  {\begingroup
   \sethlcolor{teachercol}%
   \textcolor{teachertxt}{\bfseries\hl{\ttfamily\footnotesize #1}}%
   \endgroup}%
}

\definecolor{qframe}{HTML}{0C4A6E}
\definecolor{qback}{HTML}{E0F2FE}
\definecolor{qtitlebg}{HTML}{0C4A6E}

\definecolor{rframe}{HTML}{0284C7}
\definecolor{rback}{HTML}{F0F9FF}
\definecolor{rtitlebg}{HTML}{BAE6FD}
\definecolor{rtitletxt}{HTML}{0C4A6E}

\tcbset{questionbox/.style={
    enhanced, breakable,
    colback=qback, colframe=qframe,
    boxrule=0.8pt, arc=3pt,
    left=8pt, right=8pt, top=6pt, bottom=6pt,
    fonttitle=\bfseries\small, fontupper=\small,
    coltitle=white, colbacktitle=qtitlebg,
    boxsep=2pt, before skip=10pt, after skip=10pt,
}}

\tcbset{rolloutbox/.style={
    enhanced, breakable,
    colback=rback, colframe=rframe,
    boxrule=0.6pt, arc=3pt,
    left=6pt, right=6pt, top=5pt, bottom=5pt,
    fonttitle=\bfseries\small,
    fontupper=\ttfamily\footnotesize\raggedright,
    coltitle=rtitletxt, colbacktitle=rtitlebg,
    boxsep=2pt, before skip=8pt, after skip=8pt,
}}

\subsection{Setup}

We evaluate offline how the JSD routing threshold qualitatively changes
rollouts on the 507-question SciKnowEval Chemistry-L3 multiple-choice
evaluation set. No weights are trained. Instead, at each decoding step we run
two forward passes of the same \texttt{Qwen3-4B-Instruct-2507} checkpoint: a
student pass conditioned only on the question, and a teacher pass conditioned on
the question plus privileged context containing the ground-truth answer. We then
route each generated token according to the JSD between the two next-token
distributions. This isolates how changing the routing threshold $\tau$ changes
which tokens come from the teacher and how those interventions alter the
resulting rollout. Decoding is capped at $1024$ tokens, and a response that
reaches the cap without an answer tag is scored as incorrect; this is why
the student-only accuracy here ($12.8\%$) is lower than the closed-book
base accuracy in Table~\ref{tab:retrieval_appendix}, which uses a
$16384$-token budget.

\subsection{Aggregate sweep}

Table~\ref{tab:jsd-routing-sweep} summarizes the sweep.

\begin{table}[h]
\centering
\small
\caption{
JSD-routed inference on SciKnowEval Chemistry-L3. Routed decoding recovers most
of the teacher's accuracy while using only a small fraction of teacher tokens.
}
\label{tab:jsd-routing-sweep}
\begin{tabular}{lrrr}
\toprule
\textbf{Strategy} & \textbf{Acc.} & \textbf{Mean resp.} & \textbf{Teacher frac.} \\
                  &               & \textbf{length}     & \textbf{(per-row)} \\
\midrule
Student only            & 12.8\% & 879 & \phantom{0}0.0\% \\
\ssd{}, $\tau=0.6$      & 34.12\% & 850 & \phantom{0}1.3\% \\
\ssd{}, $\tau=0.5$      & 51.87\% & 745 & \phantom{0}2.7\% \\
\ssd{}, $\tau=0.3$      & 66.07\% & 588 & \phantom{0}7.0\% \\
\ssd{}, $\tau=0.1$      & 73.96\% & 523 & 15.7\% \\
Teacher only            & 80.28\% & 487 & 100.0\% \\
\bottomrule
\end{tabular}
\end{table}

At $\tau=0.1$, only $15.7\%$ of tokens per question come from the teacher on
average, yet accuracy reaches $73.96\%$, within roughly six percentage points
of the pure-teacher ceiling. This supports the central intuition behind
\ssd{}: teacher signal is often concentrated at a small number of decisive
positions.

\subsection{Qualitative case study}

We now show a representative synthesis question.

\begin{tcolorbox}[questionbox, title={Question (SciKnowEval Chemistry-L3, reference answer: B)}]
Which of the following molecules is the only correct reactant used in the
synthesis of the product
\texttt{OCCN1CCN(CC(F)(F)F)CC1}?

\medskip
\textbf{A:} \texttt{CC(C)[Si](OCc1cncc(C(C)(C)F)c1)(C(C)C)C(C)C}

\textbf{B:} \texttt{CC(C)[Si](OCCN1CCN(CC(F)(F)F)CC1)(C(C)C)C(C)C}

\textbf{C:} \texttt{CC(C)[Si](C\#Cc1cc(Br)cc(N)n1)(C(C)C)C(C)C}

\textbf{D:} \texttt{CC(C)[Si](OCC[C@H](C)OS(C)(=O)=O)(C(C)C)C(C)C}
\end{tcolorbox}

Option B contains the exact product fragment
\texttt{OCCN1CCN(CC(F)(F)F)CC1} attached to a TIPS silyl protecting group, so
removing the silyl ether yields the product.

Table~\ref{tab:jsd-routing-case} summarizes the single-example behavior. In
the rollouts below, \T{coral-highlighted spans} were emitted from the teacher
distribution; unhighlighted text was emitted from the student distribution.
Token boundaries follow the model's BPE, and \texttt{\textbackslash n} denotes
a newline. The provenance trails come from a single-prompt deterministic rerun,
so the exact routing differs slightly from the batched aggregate sweep, but the
qualitative behavior is the same.

\begin{table}[h]
\centering
\small
\caption{
Single-question JSD-routed rollouts. In this single-prompt rerun, only
$\tau=0.1$ routes enough teacher tokens to escape the student's parsing loop
and produce the correct answer. The teacher count does not fall steadily as $\tau$
grows: each teacher token changes the text that every later step conditions on,
so the runs see different prefixes after their first switch.
}
\label{tab:jsd-routing-case}
\begin{tabular}{lrrrr}
\toprule
\textbf{Strategy} & \textbf{Teacher tokens} & \textbf{Teacher frac.} & \textbf{Length} & \textbf{Extracted} \\
\midrule
Student only            & 0   & 0.0\%   & 1024 & no tag \\
\ssd{}, $\tau=0.6$      & 11  & 1.1\%   & 1024 & no tag \\
\ssd{}, $\tau=0.5$      & 4   & 0.4\%   & 1024 & no tag \\
\ssd{}, $\tau=0.3$      & 9   & 0.9\%   & 1024 & no tag \\
\ssd{}, $\tau=0.1$      & 75  & 9.0\%   & 837  & B \\
Teacher only            & 549 & 100.0\% & 549  & B \\
\bottomrule
\end{tabular}
\end{table}

\paragraph{Student-only: recursive parsing failure.}
The student repeatedly reparses the target SMILES string and reaches the
1024-token budget without emitting an answer.

\begin{tcolorbox}[rolloutbox, title={Student-only, \textbf{incorrect}: no \texttt{<answer>} tag; length $=1024$; teacher tokens $=0$}]
\textless{}reasoning\textgreater{}\textbackslash{}n
We are given a target product: \textquotedbl{}OCCN1CCN(CC(F)(F)F)CC1\textquotedbl{}.\textbackslash{}n\textbackslash{}n
Let's first understand the structure of this molecule.\textbackslash{}n\textbackslash{}n
The product is: O--C--C--N--1--C--C--N--(C(C(F)(F)F))--C--C--1\textbackslash{}n\textbackslash{}n
More clearly, it is:\textbackslash{}n\textbackslash{}n
O--C--C--N--(CH$_2$)--N--(CH$_2$)--C(C(F)(F)F)\textbackslash{}n\textbackslash{}n
Wait --- actually, the notation \textquotedbl{}OCCN1CCN(CC(F)(F)F)CC1\textquotedbl{} is a condensed
SMILES-like string.\textbackslash{}n\textbackslash{}n
Let's parse it step by step:\textbackslash{}n\textbackslash{}n
- \textquotedbl{}OCC\textquotedbl{} $\rightarrow$ O--C--C\textbackslash{}n
- \textquotedbl{}N1\textquotedbl{} $\rightarrow$ a nitrogen atom with a ring closure at position 1
\textit{[\dots truncated]}
\end{tcolorbox}

\paragraph{Moderate thresholds: small nudges are insufficient.}
At $\tau=0.6$, $\tau=0.5$, and $\tau=0.3$, the teacher fires on a small number
of framing tokens. This changes the opening of the response, but in this
single-prompt run it does not redirect the model enough to produce an answer
before the token budget.

\begin{tcolorbox}[rolloutbox, title={JSD switch, $\tau=0.6$, \textbf{incorrect}: no \texttt{<answer>} tag; length $=1024$; teacher tokens $=11$}]
\textless{}reasoning\textgreater{}\textbackslash{}n
\T{The} target product is \textquotedbl{}OCCN1CCN(CC(F)(F)F)CC1\textquotedbl{}.\T{ This} is a molecular
formula that can be interpreted as a substituted amine structure. Let's break
it down:\textbackslash{}n\textbackslash{}n
- \textquotedbl{}OCCN1CCN(CC(F)(F)F)CC1\textquotedbl{} suggests a chain with an oxygen attached to a
carbon, followed by a nitrogen, and then a ring structure. The ring is likely a
5-membered or 6-membered heterocycle with a nitrogen and a substituent:
CC(F)(F)F, which is trifluoromethyl.
\textit{[\dots truncated]}
\end{tcolorbox}

\begin{tcolorbox}[rolloutbox, title={JSD switch, $\tau=0.5$, \textbf{incorrect}: no \texttt{<answer>} tag; length $=1024$; teacher tokens $=4$}]
\textless{}reasoning\textgreater{}\textbackslash{}n
\T{The} question asks for the only correct reactant used in the synthesis of
the product\T{ \textquotedbl{}}OCCN1CCN(CC(F)(F)F)CC1\textquotedbl{}.\textbackslash{}n\textbackslash{}n
First, let's analyze the product:\textbackslash{}n
\textquotedbl{}OCCN1CCN(CC(F)(F)F)CC1\textquotedbl{}\textbackslash{}n
This is a substituted amine structure. Breaking it down\T{:}\textbackslash{}n
- \textquotedbl{}OCC\textquotedbl{} suggests a methoxy group or --OCH$_2$--\textbackslash{}n
- \textquotedbl{}N1CCN\textquotedbl{} indicates a chain with a nitrogen atom, likely part of a cyclic
amine\textbackslash{}n
- \textquotedbl{}CC(F)(F)F\textquotedbl{} is a trifluoromethyl group
\textit{[\dots truncated]}
\end{tcolorbox}

\begin{tcolorbox}[rolloutbox, title={JSD switch, $\tau=0.3$, \textbf{incorrect}: no \texttt{<answer>} tag; length $=1024$; teacher tokens $=9$}]
\textless{}reasoning\textgreater{}\textbackslash{}n
\T{The} question asks for the only correct reactant used in the synthesis of
the product\T{ \textquotedbl{}}OCCN1CCN(CC(F)(F)F)CC1\textquotedbl{}.\textbackslash{}n\textbackslash{}n
First, let's analyze the product:\T{ \textquotedbl{}}OCCN1CCN(CC(F)(F)F)CC1\T{\textquotedbl{}.}
This is a substituted amine structure.\T{ The} core structure appears to be:
\textbackslash{}n\textbackslash{}n
- O--\T{CH}$_2$--N--CH$_2$--N--(CH$_2$)--CC(F)(F)F
\textbackslash{}n\textbackslash{}n
More precisely, it's O--CH$_2$--N--CH$_2$--N--(CH$_2$)--C(F)(F)--CH? Wait
--- let's parse it properly.
\textit{[\dots truncated]}
\end{tcolorbox}

\paragraph{Low threshold: routing changes the strategy.}
At $\tau=0.1$, teacher routing covers a denser set of setup tokens. The rollout
moves from recursive SMILES parsing to the correct structural-motif strategy:
identify which answer choice contains the product fragment.

\begin{tcolorbox}[rolloutbox, title={JSD switch, $\tau=0.1$, \textbf{correct}: extracted B; length $=837$; teacher tokens $=75$}]
\textless{}reasoning\textgreater{}\textbackslash{}n
\T{The question} asks\T{ which molecule} is the\T{ correct} reactant\T{ in}
the synthesis of the product\T{ \textquotedbl{}}OCCN1CCN(CC(F)(F)F)CC1\textquotedbl{}.
\textbackslash{}n\textbackslash{}n
First,\T{ analyze} the product:\T{ \textquotedbl{}}OCCN1CCN(CC(F)(F)F)CC1\textquotedbl{} is\T{}
a\T{ substituted am}ine\T{ structure. It contains a} chain\T{ with} a nitrogen
atom\T{ and} a group --CC(F)(F)F, which is a trifluoromethyl group\T{ (}CF$_3$).
\T{ The} structure\T{ suggests} a\T{ diam}ine or amine derivative with a CF$_3$
group.\textbackslash{}n\textbackslash{}n
\T{Now}, look at the options\T{. The key} is to identify which reactant
contains the same\T{ core structure} as the product\T{,} especially the
\T{ \textquotedbl{}}OCCN1CCN(CC(F)(F)F)CC1\textquotedbl{} fragment.\textbackslash{}n\textbackslash{}n
Option A: Contains\T{ a phen}yl\T{ ring with a fluor}inated\T{ substit}uent
\T{, not} matching the amine
\textit{[\dots truncated]}
\end{tcolorbox}

\paragraph{Teacher-only.}
The teacher-only rollout is correct and shorter. Since every token is generated
from the privileged-context distribution, we do not highlight each token
individually.

\begin{tcolorbox}[rolloutbox, title={Teacher-only, \textbf{correct}: extracted B; length $=549$; teacher tokens $=549$}]
\emph{All tokens in this rollout are teacher-generated.}

\medskip

\textless{}reasoning\textgreater{}\textbackslash{}n
The question asks which molecule is the correct reactant in the synthesis of
the product \textquotedbl{}OCCN1CCN(CC(F)(F)F)CC1\textquotedbl{}.\textbackslash{}n\textbackslash{}n
First, analyze the product: \textquotedbl{}OCCN1CCN(CC(F)(F)F)CC1\textquotedbl{} is a substituted amine
structure. It contains a chain with a nitrogen atom and a group CC(F)(F)F,
which is a trifluoromethyl group (CF$_3$). The structure suggests a diamine or
amine derivative with a CF$_3$ group.\textbackslash{}n\textbackslash{}n
Now, look at the options. The key is to find which reactant contains the exact
same functional group as the product, specifically the
\textquotedbl{}OCCN1CCN(CC(F)(F)F)CC1\textquotedbl{} fragment.
\textit{[\dots truncated]}
\end{tcolorbox}

The visible pattern is that teacher routing fires on framing and setup tokens,
not on the answer letter itself. The student-only rollout opens by trying to
parse the SMILES string and gets stuck in a loop. Moderate thresholds insert a
few teacher tokens, such as ``The'', quote marks, and small structural tokens,
which weakly shift the framing but do not change the outcome. At $\tau=0.1$,
teacher routing covers enough of the setup to move the rollout into the
correct strategy: analyze the product, identify the shared structural motif,
and choose the option containing that motif.